\documentclass{article}
\pdfoutput=1
\usepackage[final]{neurips_2021}
\usepackage[utf8]{inputenc} % allow utf-8 input
\usepackage[T1]{fontenc}    % use 8-bit T1 fonts
\usepackage{hyperref}       % hyperlinks
\usepackage{url}            % simple URL typesetting
\usepackage{booktabs}       % professional-quality tables
\usepackage{amsfonts}       % blackboard math symbols
\usepackage{nicefrac}       % compact symbols for 1/2, etc.
\usepackage{microtype}      % microtypography
\usepackage{xcolor}         % colors

\usepackage{url}
\usepackage{amsmath}
\usepackage{graphicx}
\usepackage{xspace}
\usepackage{multirow}
\usepackage{tabularx}
\usepackage[export]{adjustbox}
\usepackage{chngpage}
\usepackage{bm}
\usepackage{dsfont}
\usepackage{physics}
\usepackage{amssymb}
\usepackage{booktabs}
\usepackage{enumitem}
\usepackage{wrapfig}
\usepackage{colortbl}
\usepackage{pifont}
\usepackage{listings}
\usepackage{algorithm}
\usepackage{algpseudocode}
\usepackage{booktabs}
\usepackage{epstopdf}
\usepackage{amssymb}
\usepackage{pdfpages}
\usepackage{pifont}
\usepackage{subcaption}
\usepackage{adjustbox}
\usepackage{fancyhdr}
\usepackage{tikz}
\usepackage{float}
\usepackage{wrapfig}

\def\ourmethod{{GR-2}\xspace}

\title{
\ourmethod: A Generative Video-Language-Action Model with Web-Scale Knowledge for Robot Manipulation
}

\author{%
\thanks{Authors are listed in alphabetical order. Contributions are listed at the end of the report. Corresponding email(s): \texttt{\{kongtao,wuhongtao.123\}@bytedance.com}}~~Chi-Lam Cheang, ~~~~Guangzeng Chen, ~~~~Ya Jing, ~~~~Tao Kong, \\ \\
~~~~\textbf{Hang Li}, ~~~~\textbf{Yifeng Li}, ~~~~\textbf{Yuxiao Liu,} ~~\textbf{Hongtao Wu,} \\ \\
~~~~\textbf{Jiafeng Xu,} ~~~~\textbf{Yichu Yang,}  ~~~~\textbf{Hanbo Zhang,} ~~~~\textbf{Minzhao Zhu}
\\ \\ 
ByteDance Research
}

\begin{document}
\maketitle

\begin{abstract}
We present \ourmethod, a state-of-the-art generalist robot agent for versatile and generalizable robot manipulation.
\ourmethod is first pre-trained on a vast number of Internet videos to capture the dynamics of the world.
This large-scale pre-training, 
involving 38 million video clips and over 50 billion tokens, 
equips \ourmethod with the ability to generalize across a wide range of robotic tasks and environments during subsequent policy learning.
Following this, \ourmethod is fine-tuned for both video generation and action prediction using robot trajectories. It exhibits impressive multi-task learning capabilities, achieving an average success rate of 97.7\% across more than 100 tasks. 
Moreover, \ourmethod demonstrates exceptional generalization to new, previously unseen scenarios, including novel backgrounds, environments, objects, and tasks. 
Notably, \ourmethod scales effectively with model size, underscoring its potential for continued growth and application. Project page: \url{https://gr2-manipulation.github.io}.
\end{abstract}

\section{Introduction}
The rise of high-capacity foundation models has contributed significantly to the success of language~\cite{achiam2023gpt}, image~\cite{ravi2024sam}, and video~\cite{videoworldsimulators2024} processing tasks. 
These models are initially pre-trained on large-scale diverse datasets and can subsequently be adapted to specific downstream tasks, making them versatile in application.
This paradigm allows these models to tackle a variety of tasks with a single generalist model when conditioned on different inputs (\textit{e.g.}, language prompts~\cite{brown2020language}).

Following the foundation models established in other domains, our goal is to develop a foundation generalist manipulation agent via large-scale pre-training on a comprehensive dataset. 
This would enable rapid adaptation to a wide range of novel manipulation tasks via efficient fine-tuning.
A generalist manipulation agent should be capable of executing a wide range of manipulation skills.
And more importantly, it should exhibit strong performance in acquiring new skills and handling disturbances.
Despite recent advances in AI and a shift towards data-driven learning, collecting large-scale robot data remains a significant challenge due to inefficient data collection methods and the limited scalability of real-robot systems.
Research suggests that pre-training on video generation can effectively transfer valuable knowledge from videos to policy learning, thus improving the action prediction capability~\cite{wu2023unleashing}.

This report introduces \ourmethod, an evolution of our previous model~\cite{wu2023unleashing}, featuring improved performance and expanded capabilities. 
To achieve this, we pre-train \ourmethod on an extensive video dataset encompassing diverse daily human activities across different contexts (household, outdoor, workplace, leisure, etc.).
The primary pre-training objective is straightforward: given a textual description and a video frame, the model predicts subsequent frames based on the text.
By mastering this auto-regressive prediction task, we anticipate the model to capture crucial temporal dynamics and semantic information which are essential for downstream policy learning.
Through fine-tuning on robot trajectories, \ourmethod demonstrates the capability to learn multiple manipulation tasks and adapt to novel scenarios, including novel backgrounds, environments, objects, and tasks.
Notably, \ourmethod efficiently learns over 100 tasks from a dataset with only 5,000 trajectories (an average of 50 trajectories per task).
This significantly reduces the cost of acquiring new skills and adapting to new environments in application.
Furthermore, \ourmethod excels in generalizing to unseen objects in an end-to-end bin-picking setting, highlighting its strong potential for industrial application.
Specifically, \ourmethod builds upon GR-1~\cite{wu2023unleashing} with several key improvements:
\begin{enumerate}[leftmargin=*]
    \item[$\bullet$] \ourmethod is pre-trained on 38 million text-video data (amounting to over 50 billion tokens), and is capable of accomplishing over 100 manipulation tasks and performing bin-picking of over 100 objects. It significantly scales up the pre-training data and number of tasks.
    \item[$\bullet$] We develop a novel model architecture that allows the knowledge gathered from pre-training to seamlessly transfer to downstream fine-tuning in a lossless way. The model demonstrates strong scalability in handling multiple tasks in challenging generalization settings.
    \item[$\bullet$] For real-robot deployment, we introduce a whole-body control (WBC) algorithm that incorporates trajectory optimization and real-time motion tracking.
\end{enumerate}

The remainder of this report is organized as follows.
Sec.~\ref{sec:method} provides a detailed description of \ourmethod, including its model architecture, training process, and real-world deployment.
Sec.~\ref{sec:exp} outlines our experiment setups and results.
Sec.~\ref{sec:related_work} discusses the relation of \ourmethod to existing works.
Finally, Sec.~\ref{sec:conclusion} concludes the work and discusses future directions.

\begin{figure}[t]
    \centering
    \includegraphics[width=0.95\textwidth]{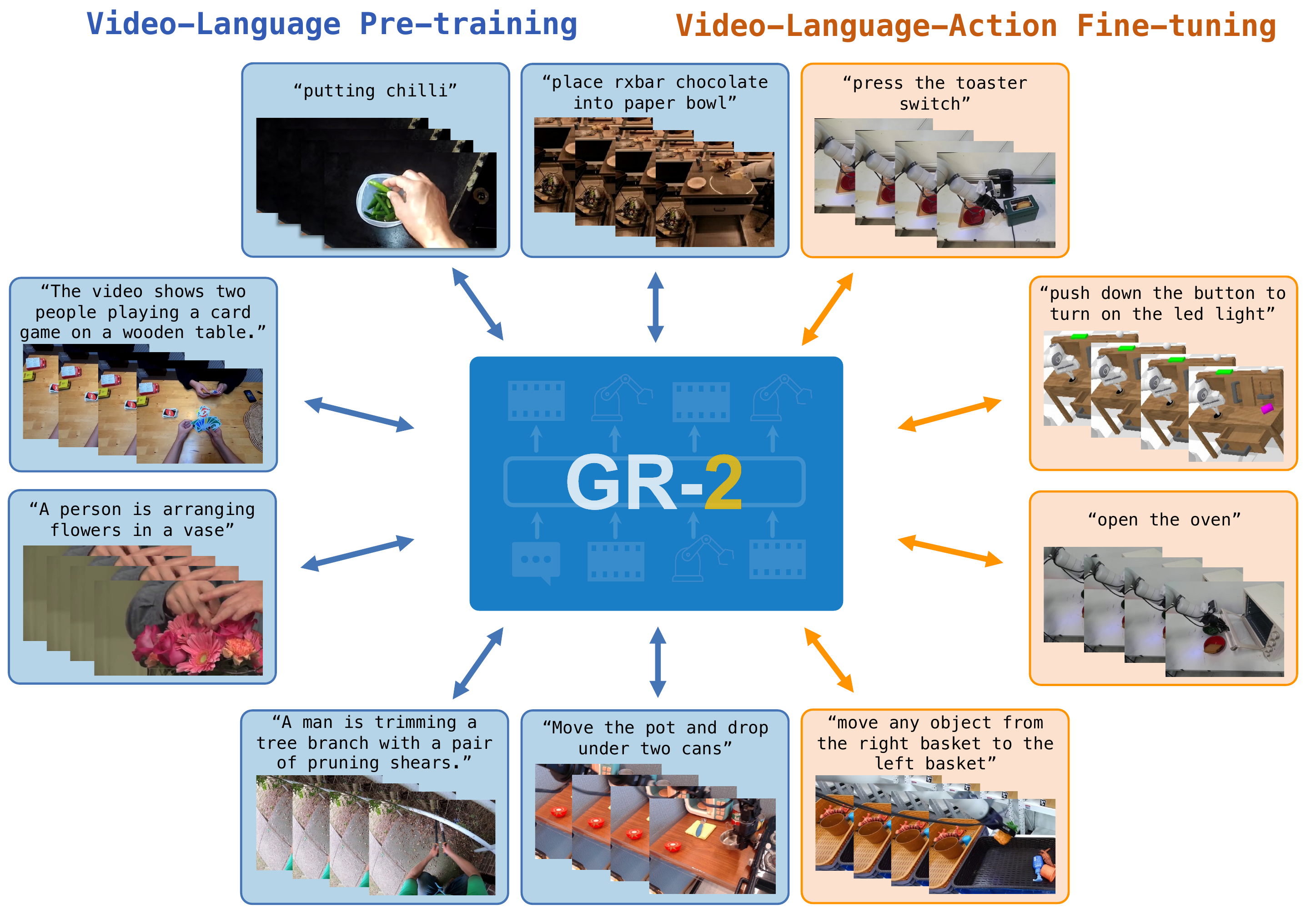}
    \caption{
        \textbf{Overview.}
        \ourmethod undegoes two stages of training: video generation pre-training and robot data fine-tuning.
    }
    \label{fig:overview}
\end{figure}

\section{Methods}
\label{sec:method}
We consider language-conditioned visual robot manipulation as our approach towards generalist robot manipulation, as language is one of the most flexible ways for a human to specify tasks for a robot.
In this setting, a single robot policy must solve multiple complex manipulation tasks by understanding different unconstrained language instructions.
Specifically, we want to train a universal policy $\pi$ that takes as inputs a language instruction $l$, a sequence of environment observation $\mathbf{o}_{t-h:t}$, and a sequence of robot states $\mathbf{s}_{t-h:t}$.
The policy outputs an action trajectory $\mathbf{a}_{t:t+k}$ in an end-to-end manner:
\begin{equation}
    \mathbf{a}_{t:t+k} = \pi(l, \mathbf{o}_{t-h:t}, \mathbf{s}_{t-h:t}),
\end{equation}
where $h$ and $k$ denote the length of the observation history and the action trajectory, respectively.

\subsection{Model \& Training}
\ourmethod is a language-conditioned GPT-style visual manipulation policy model (Fig.~\ref{fig:overview}).
The training undergoes two stages: video generative pre-training and robot data fine-tuning.
During the pre-training stage, we train \ourmethod on a curated large-scale video dataset.
After that, we fine-tune \ourmethod on robot data to predict action trajectories \textit{and} videos in tandem:
\begin{equation}
\pi(l, \mathbf{o}_{t-h:t}, \mathbf{s}_{t-h:t}) \rightarrow \mathbf{o}_{t+1}, \mathbf{a}_{t:t+k}
\end{equation}
The inputs to \ourmethod contain a language instruction, a sequence of video frames, and a sequence of robot states.

We use a frozen text encoder~\cite{radford2021learning} to tokenize the language instruction.
For the image frames in the video, we employ a VQGAN~\cite{esser2021taming} to convert each image into discrete tokens.
The VQGAN is trained on a large corpus of Internet data as well as in-domain robot data and is kept frozen during the training process. This approach facilitates fast training and supports the generation of high-quality videos.
Robot states contain the position and rotation of the end-effector, as well as the binary gripper state.
The states are encoded via linear layers, which are trainable during the fine-tuning stage.

\begin{figure}[t]
    \centering
    \includegraphics[width=\columnwidth]{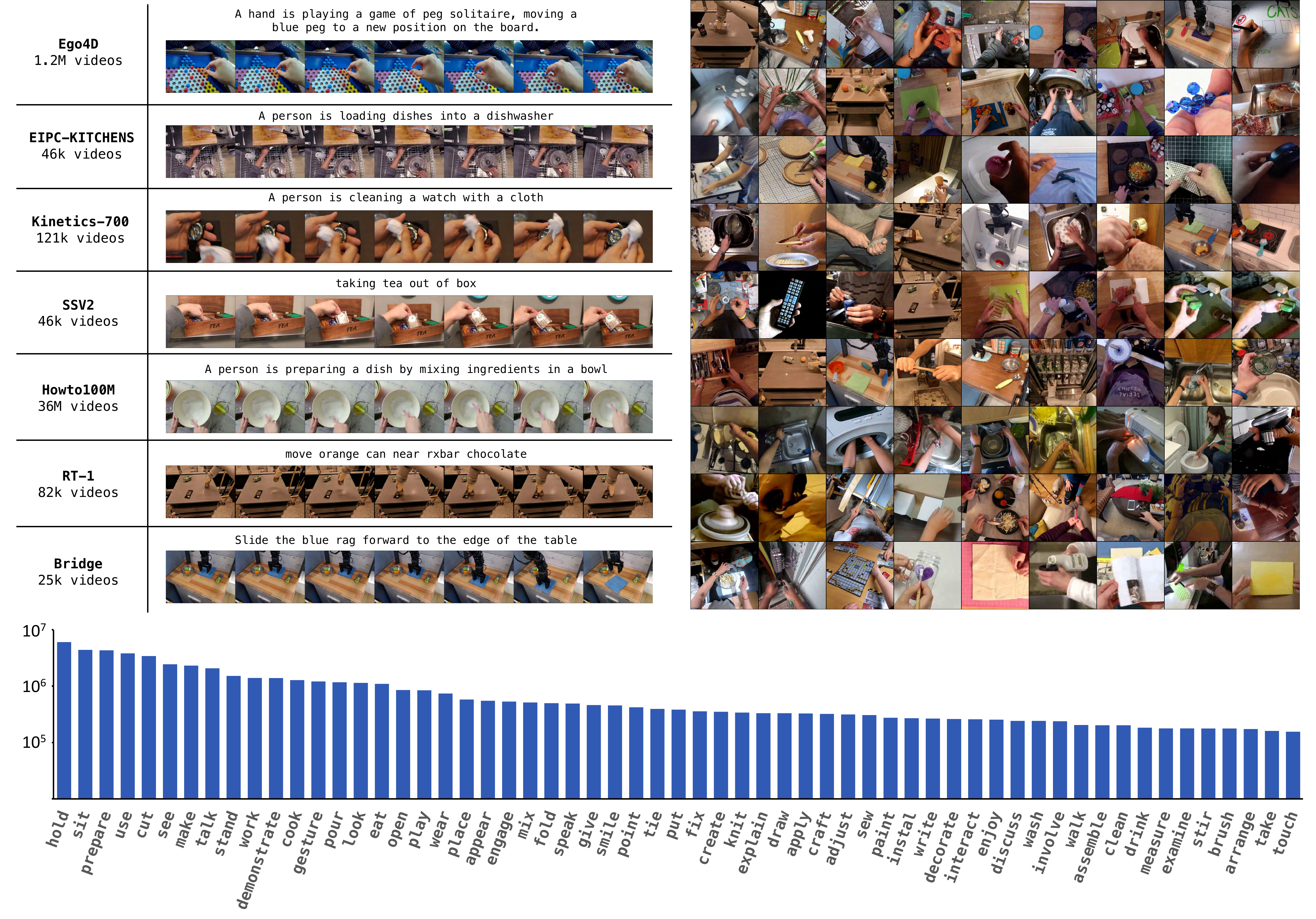}
    \caption{
        \textbf{Pre-training Dataset.}
        We show sample videos and the verb distribution of the pre-training dataset we curated. 
        The y-axis of the bottom plot is the logarithm frequency of the top words.
    }
    \label{fig:pt_dataset}
\end{figure}

Our goal in the pre-training stage is to equip \ourmethod with the capability to predict future videos. This enables the model to develop a strong prior for predicting future events, thereby enhancing its ability to make accurate action predictions.
The model, built upon a GPT-style transformer, takes the tokenized text and image sequence as inputs and outputs the discrete tokens of future images.
Future images are decoded from these tokens with the VQGAN decoder.
We highlight that \ourmethod is pre-trained on a significantly larger volume of video data compared to previous works that utilize video pre-training.
The pre-training data includes commonly used public datasets of human activities, \textit{e.g.}, Howto100M~\cite{miech2019howto100m}, Ego4D~\cite{grauman2022ego4d}, Something-Something V2~\cite{goyal2017something}, EPIC-KITCHENS~\cite{damen2018scaling}, and Kinetics-700~\cite{carreira2019short}. 
To tailor the pre-training data for robot manipulation tasks, we carefully establish a data processing pipeline that includes hand filtering~\cite{lugaresi2019mediapipe} and re-captioning~\cite{opensora}.
In addition, we include publicly available robot datasets, \textit{e.g.}, RT-1~\cite{brohan2022rt} and Bridge~\cite{walke2023bridgedata}.
In total, the number of video clips used for pre-training is 38 million, equivalent to approximately 50 billion tokens. 
The distribution of human activities and video samples are illustrated in Fig.~\ref{fig:pt_dataset}.

\ourmethod can be seamlessly fine-tuned on robot data after large-scale pre-training.
Unlike the videos in pre-training data which only have a single camera view, robot data usually contain multiple views.
\ourmethod is designed to gracefully handle multiple views. 
It takes as inputs the tokenized language instruction, the image sequences captured from multiple views, and the robot state sequence.
The outputs include future images of each view and an action trajectory.
The action trajectory is generated with a conditional VAE (cVAE)~\cite{sohn2015learning, kingma2013auto,zhao2023learning}. 
We empirically found that generating action trajectories rather than single-step actions is crucial for both trajectory smoothing and real-time performance.

\subsection{Real-Robot System \& Deployment}
Our real-robot system consists of a 7-DoF Kinova Gen3 robot arm paired with a Robotiq 2F-85 gripper. 
We utilize two cameras: a static head camera provides an overview of the workspace; another camera, which is mounted on the end-effector, offers a close-up view of interactions between the gripper and the environment.

\ourmethod generates an action trajectory in Cartesian space. 
To ensure that the robot arm accurately follows this trajectory, we develop a Whole-Body Control (WBC) algorithm that employs trajectory optimization for motion tracking~\cite{yang2023moma-force}. 
The generated trajectory first undergoes optimization to improve its smoothness and continuity.
Subsequently, the WBC algorithm converts the Cartesian trajectory into low-level joint actions, which are executed on the real robot at a frequency of 200 Hz.
This process integrates collision constraints and manipulability into the optimization framework. 

\section{Experiments}
\label{sec:exp}
\begin{figure}
    \centering
    \includegraphics[width=\columnwidth]{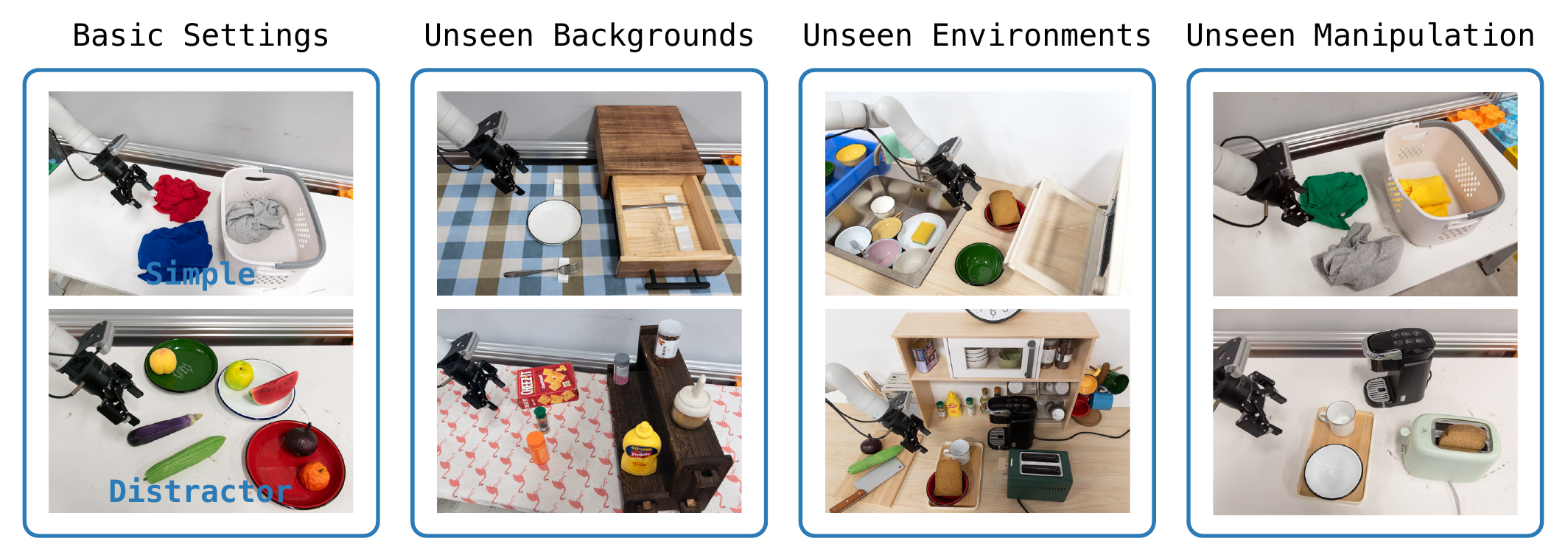}
    \caption{\textbf{Multi-Task Learning.} We perform experiments in two basic settings (Simple and Distractor) and three generalization settings (Unseen Backgrounds, Unseen Environments, and Unseen Manipulation).}
    \label{fig:multi_task_setting}
\end{figure}

\begin{figure}[htbp]
    \centering
    \includegraphics[width=\columnwidth, center]{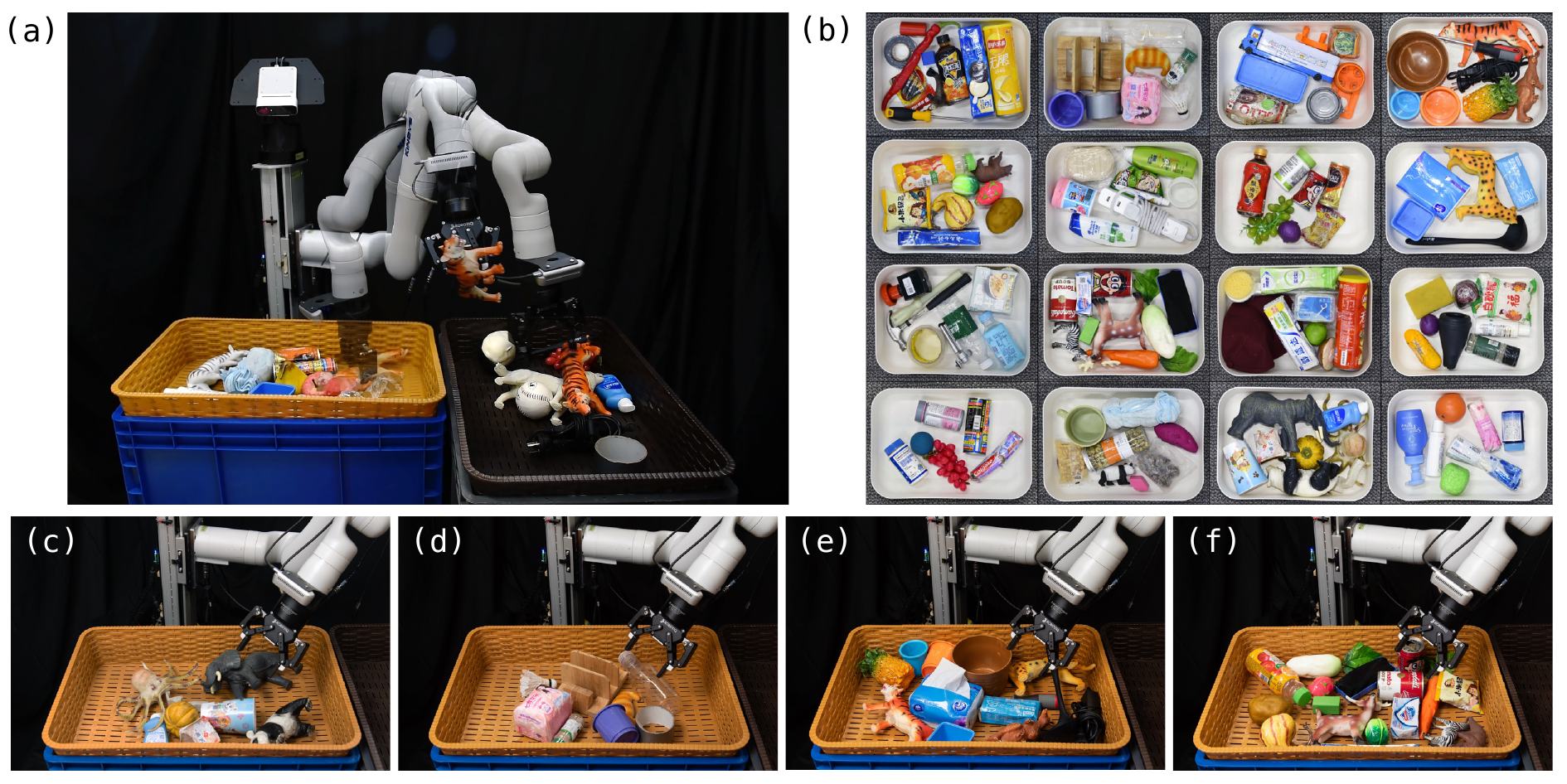}
    \caption{
        \textbf{End-to-End Bin Picking.}
        (a) Experiment setting.
        (b) Objects used in the experiments. We evaluate in four different settings:
        (c) Seen,
        (d) Unseen,
        (e) Cluttered Seen, and 
        (f) Cluttered Unseen.
        Seen (Unseen) indicates that the objects are seen (unseen) during training.
        The two cluttered settings (e) and (f) have more objects compared to the training setting. 
    }
    \label{fig:bin_picking_setting}
\end{figure}

We perform large-scale real-robot experiments in two settings: multi-task learning (Fig.~\ref{fig:multi_task_setting}) and end-to-end bin picking (Fig.~\ref{fig:bin_picking_setting}).
In multi-task learning, we aim to evaluate the capability of \ourmethod on learning multiple different tasks. 
We also evaluate in multiple challenging out-of-distribution settings to verify its generalization capabilities (Fig.~\ref{fig:multi_task_setting}).
In end-to-end bin picking, our goal is to evaluate \ourmethod in a more industrial setting. 
In this setting, the model is provided with a single text prompt and is required to perform the bin-picking task within an object cluster.
Finally, we present a benchmark comparison with state-of-the-art methods on the challenging CALVIN benchmark~\cite{mees2022calvin}. 
If not specified otherwise, the default GR-2 model contains 230M parameters, of which 95M are trainable.
We also show model scaling results in Sec~\ref{sec:exp:scaling}.

\begin{figure}[t]
    \centering
    \includegraphics[width=0.98\columnwidth]{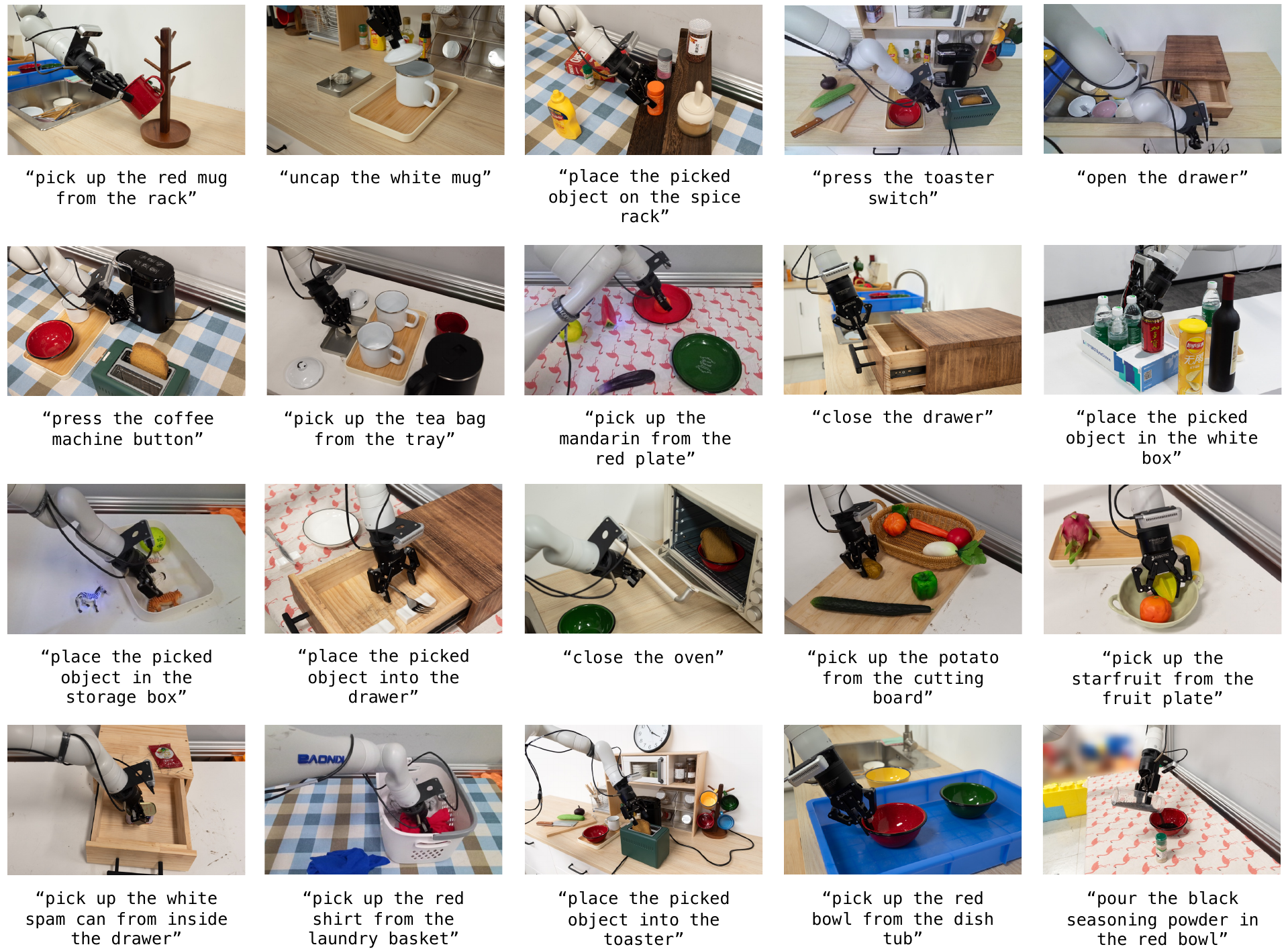}
    \caption{
    \textbf{Task Examples.}
    \ourmethod is able to perform more than 100 tasks of 8 skills including picking, placing, uncapping, capping, opening, closing, pressing, and pouring.
    }
    \label{fig:multi_task}
\end{figure}

\subsection{Real-World Multi-Task Learning}
\label{sec:exp:multi-task}
We collected human demonstrations of 105 table-top tasks via teleoperation.
These tasks cover 8 different skills, \textit{i.e.}, picking, placing, uncapping, capping, opening, closing, pressing, and pouring (Fig.~\ref{fig:multi_task}).
In total, we collected about 40,000 trajectories, with an average of 400 trajectories per task.
Based on the model pre-trained on the curated large-scale video dataset, we further fine-tune \ourmethod using this dataset. Additionally, to evaluate the performance under the condition of data scarcity, we train GR-2 using approximately 1/8 of the full dataset, which corresponds to around 50 trajectories per task.

To enable better generalization to unseen scenarios, we perform data augmentation during fine-tuning by adding new objects into the scene and/or changing the background. 
To insert new objects into the scene, a diffusion model~\cite{ho2020denoising} is trained with a combination of a self-collected object dataset and the Open Images dataset~\cite{kuznetsova2020open}. 
This model enables us to insert a specific object in a designated region. 
For changing the background, we utilize Segment Anything Model (SAM)~\cite{kirillov2023segment} to extract regions corresponding to the background. 
Finally, we employ a video generation model~\cite{ma2024latte} that conditions on the original video and the inpainted frame to produce an augmented video while preserving the robot motion.

\begin{figure}[htbp]
    \centering
    \includegraphics[width=1.0\columnwidth, center, page=1]{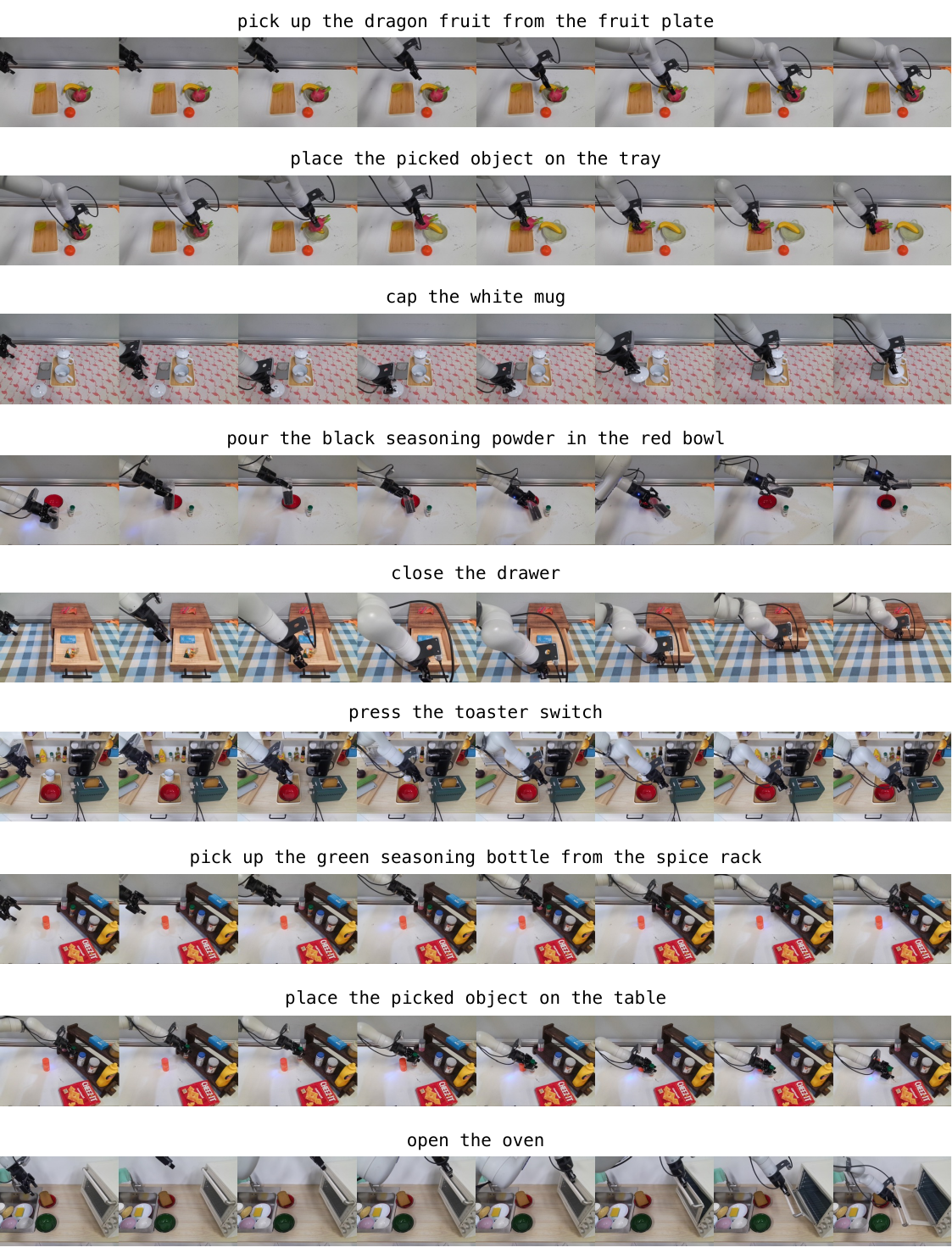}
    \caption{
        \textbf{Qualitative Results of Multi-Task Learning.}
        We show end-to-end rollouts of different tasks.
    }
    \label{fig:multi_task_rollout}
\end{figure}

\textbf{Basic Settings.}
We first evaluate \ourmethod in two basic settings: Simple and Distractor.
In Simple, the test environment is set similar to the training data.
In Distractor, we add a few distractors to the scene.
This becomes challenging for the reason that 1) distractors, especially those that share a similar color and/or shape with the target object, may confuse the robot and 2) the environment becomes more cluttered and sometimes requires collision avoidance to accomplish a task.

\begin{figure}[t]
    \centering
    \includegraphics[width=\columnwidth, center, page=1]{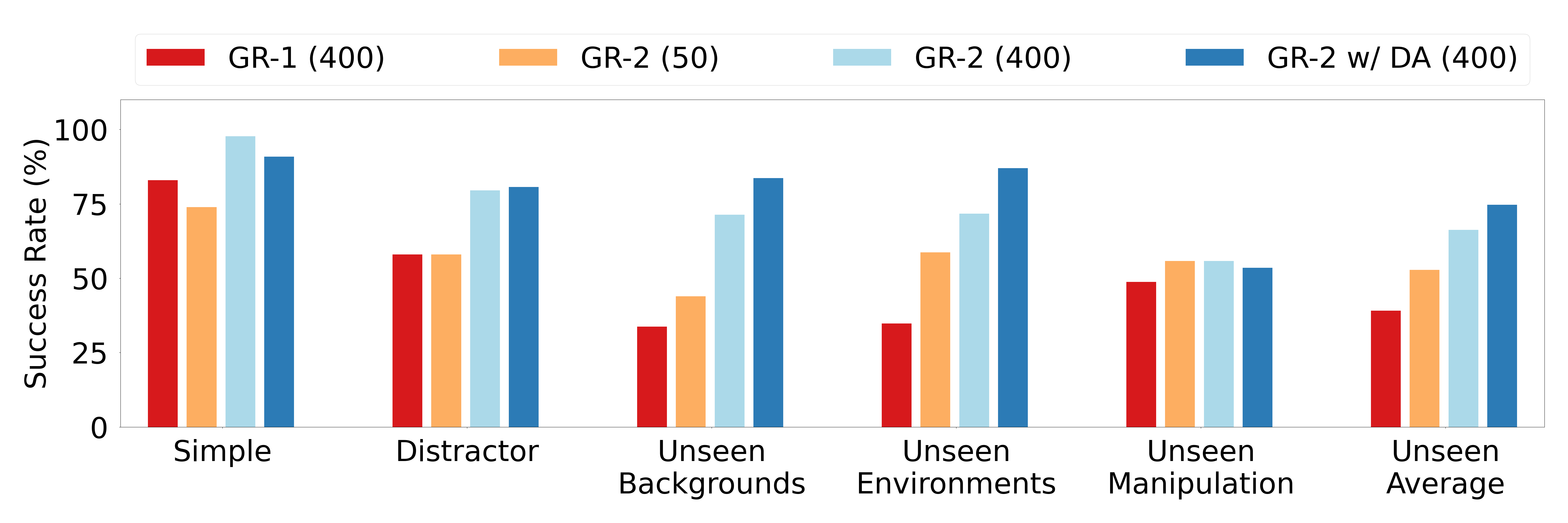}
    \caption{
        \textbf{Success Rates of Multi-Task Learning.}
        We show the success rates of four models across different evaluation settings.
        400 (50) indicates that the model is trained on about 400 (50) trajectories per task on average.
        \ourmethod w/ DA indicates that we perform data augmentation on the training data.
        See Sec.~\ref{sec:exp:multi-task} for more details.
    }
    \label{fig:multi_task_success_rate}
\end{figure}

\textbf{Generalization Settings.}
To further investigate the capability of \ourmethod in unseen scenarios, we introduce three more challenging settings: Unseen Backgrounds, Unseen Environments, and Unseen Manipulation (Fig.~\ref{fig:multi_task_setting}).
In Unseen Backgrounds, we change the background by adding two unseen tablecloths that are very different from the original plain background in the training dataset as shown in Fig~\ref{fig:multi_task_setting}.
For Unseen Environments, we evaluate in two unseen kitchen environments.
Besides changed backgrounds, these environments also contain scene-related distractors.
Finally, for Unseen Manipulation, we instruct the robot to perform manipulations that are unseen in the robot training data.
It includes manipulating objects of unseen categories and unseen object instances.
This setting is extremely challenging given the robot has never seen these objects in the training data.
And the unseen instructions for manipulating objects of unseen categories further increase the difficulties.

\textbf{Results.}
Qualitative results are shown in Fig.~\ref{fig:multi_task_rollout}.
Quantitative results are shown in Fig.~\ref{fig:multi_task_success_rate}.
\ourmethod achieves a success rate of 97.7\% on 105 tasks in the Simple setting, showcasing its powerful multi-task learning capability.
It can also robustly handle distractors and attend to target objects correctly.
It improves the success rates of GR-1 in all settings.
Notably, it achieves success rates of 71.4\% and 71.7\% in Unseen Backgrounds and Unseen Environments, respectively, doubling those of GR-1.
By introducing data augmentation, \ourmethod w/ DA is able to achieve even more competitive generalization performance, obtaining a success rate of 87.0\% in Unseen Environments and an average success rate of 74.7\% across all three generalization settings.
When trained with only 50 trajectories per task, \ourmethod is able to achieve a success rate of 73.9\% in the Simple setting and outperforms GR-1 in all three generalization settings.
This showcases the strong potential of \ourmethod in efficiently adapting to new tasks and environments.
Finally, \ourmethod achieves a success rate of 55.8\% in Unseen Manipulation. 
Typical failure cases include 1) failing to pick unseen objects of novel shapes and 2) mistakenly selecting the wrong object when instructed to pick an unseen one. 
Moving forward, we plan to explore techniques to further improve generalization for unseen manipulation tasks, including handling novel objects and executing new skills.

\subsection{End-to-End Bin Picking of Different Objects}
To further assess the capabilities of \ourmethod in an industrial context, we conduct large-scale experiments on end-to-end bin picking.
The experiment setup contains a source and a target basket (Fig.~\ref{fig:bin_picking_setting}(a)).
The robot is tasked with picking objects from the source basket and placing them into the target basket in a seamless and end-to-end manner.
In total, we collected about 94,000 pick-and-place trajectories of 55 objects for training. 
The language instruction is very simple: 
\begin{center}
    \texttt{move any object from the right basket to the left basket.} 
\end{center}

\textbf{Settings.}
We evaluate \ourmethod in four different settings: Seen, Unseen, Cluttered Seen, and Cluttered Unseen (Fig.~\ref{fig:bin_picking_setting}(c)(d)(e)(f)).
In total, we perform experiments on 122 objects, among which 55 of them are seen and the other 67 are unseen during training (Fig.~\ref{fig:bin_picking_setting}(b)).
We transport 5-9 seen (unseen) objects from the source basket to the target one in the Seen (Unseen) setting.
The number of objects in the source basket at the beginning is similar to those in the training data.
For the Cluttered Seen (Unseen) setting, we increase the number of objects by twofold, \textit{i.e.}, including 12-17 seen (unseen) objects in the source basket.
And thus the two cluttered settings can be considered as unseen settings regardless of whether the objects are seen or unseen.

\begin{figure}
    \centering
    \includegraphics[width=1.0\columnwidth]{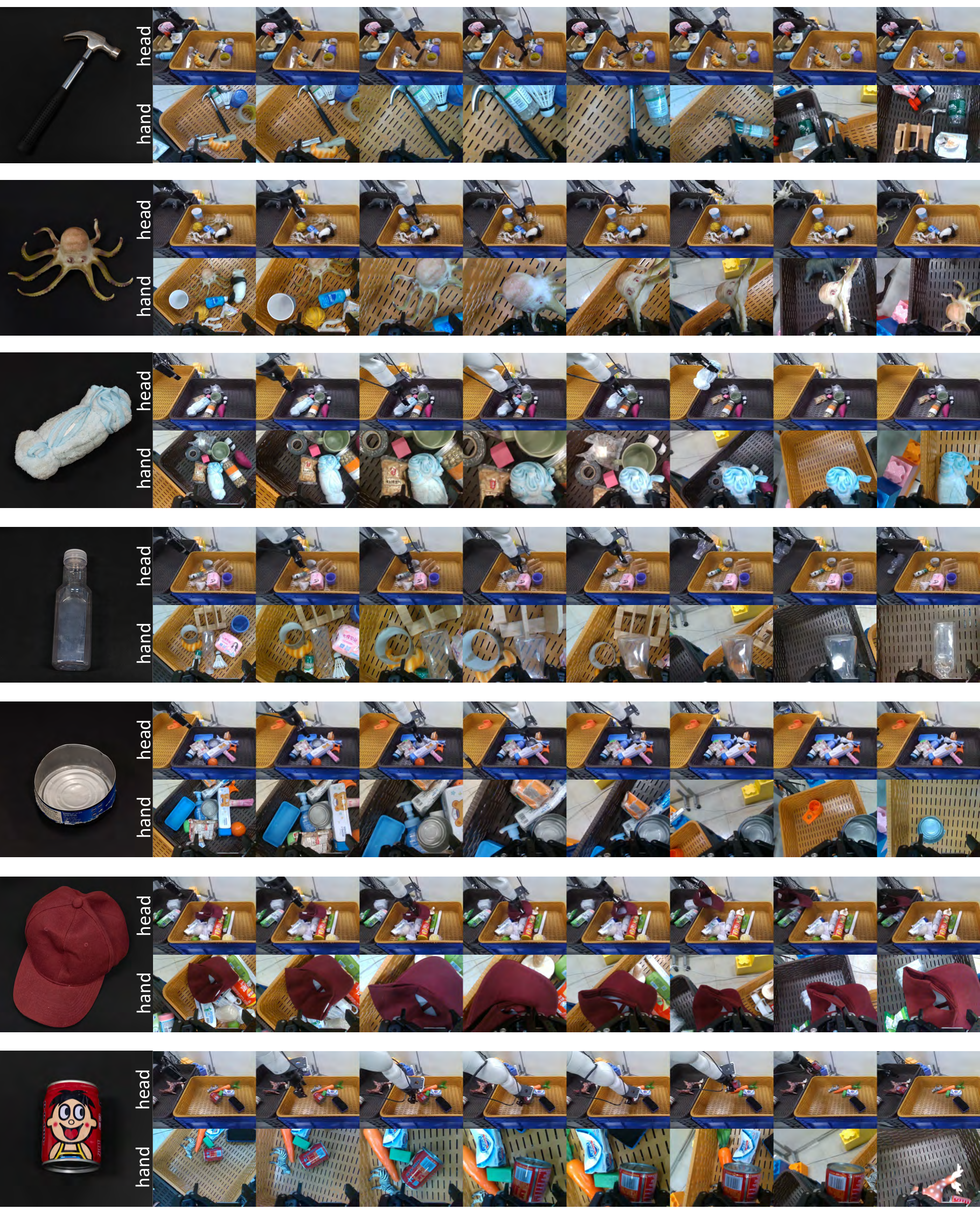}
    \caption{
    \textbf{Qualitative Results of End-to-End Bin Picking.}
    We show end-to-end picking of different objects, including objects that are transparent, deformable, or reflective.
    }
    \label{fig:bin_picking_rollout}
\end{figure}

\begin{wrapfigure}{r}{7cm}
    \centering
    \vspace{-0.6cm}
    \includegraphics[height=5cm,width=7cm]{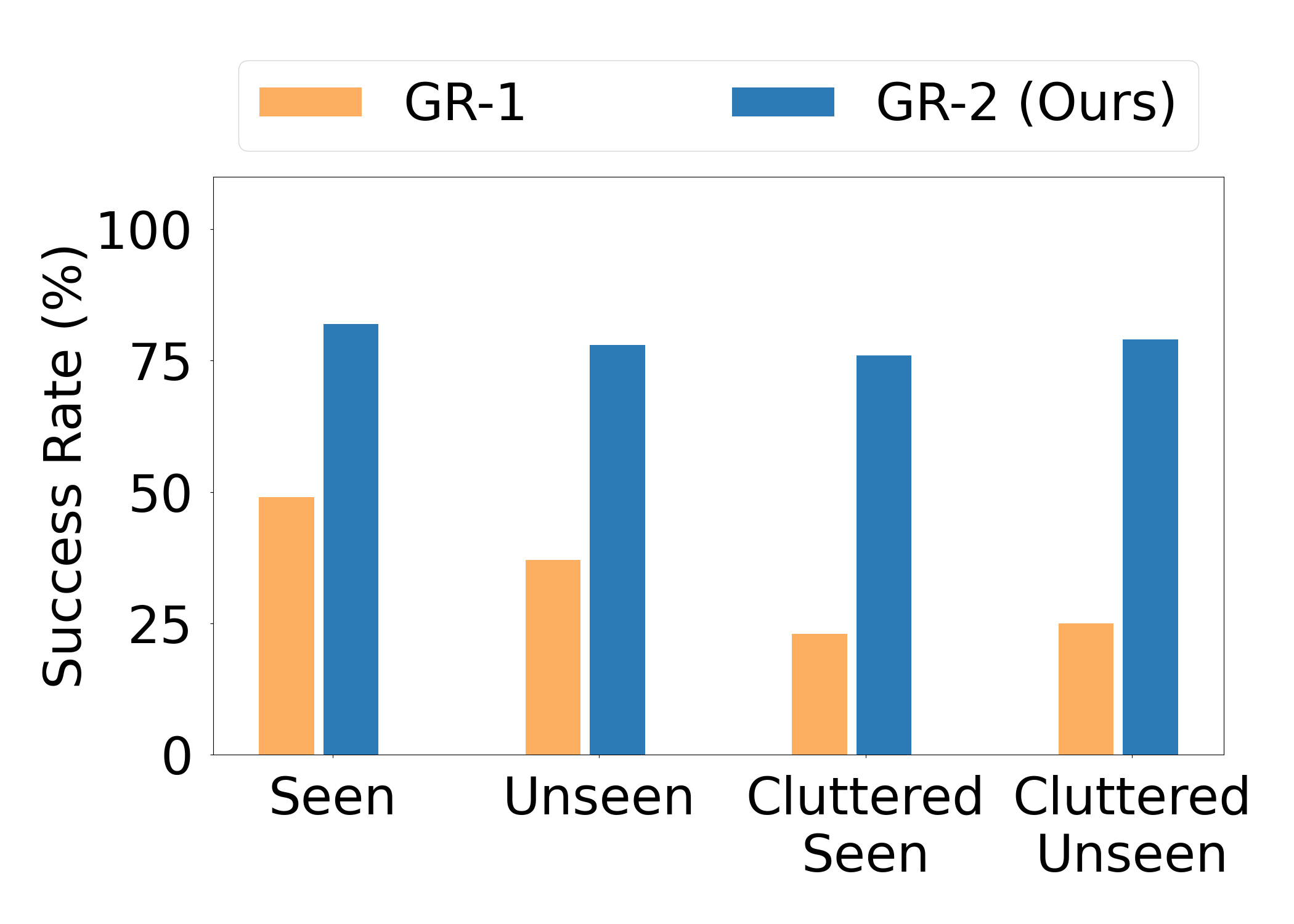}
    \vspace{-0.6cm}
    \caption{
        \textbf{Success Rates of End-to-End Bin Picking.}
    }
    \vspace{-0.5cm}
    \label{fig:bin_picking_success_rate}
\end{wrapfigure}

\textbf{Results.}
Qualitative results are shown in Fig.~\ref{fig:bin_picking_rollout}.
Quantitative results are shown in Fig.~\ref{fig:bin_picking_success_rate}.
\ourmethod outperforms GR-1 by a large margin, improving the average success rate from 33.3\% to 79.0\%.
GR-1 is not able to handle the Unseen and the two cluttered settings.
The performance degrades largely from that of the Seen setting.
On the other hand, we highlight that the success rates of \ourmethod in the Unseen and the two cluttered settings are comparable to that of the Seen setting.
These results showcase that \ourmethod possesses powerful generalization capabilities for unseen objects and unseen scenarios, indicating great potential for industrial application.
\ourmethod is able to handle objects that may be challenging for model-based methods, including transparent, deformable, and reflective objects.
See Fig.~\ref{fig:bin_picking_rollout} for some examples.

\subsection{CALVIN Benchmark}
CALVIN is a simulated benchmark which targets long-horizon language-conditioned robot manipulation~\cite{mees2022calvin}.
It includes 34 tasks and incorporates unconstrained language instructions.
We perform experiments on the ABCD-D split which includes more than 20,000 expert demonstrations of 34 different manipulation tasks.
Following~\cite{mees2022calvin}, we perform evaluation on 1,000 unique sequences of instruction chains.
For each sequence, \ourmethod is instructed to perform 5 tasks in a row.

\begin{figure}[ht]
    \centering
    \includegraphics[width=\columnwidth]{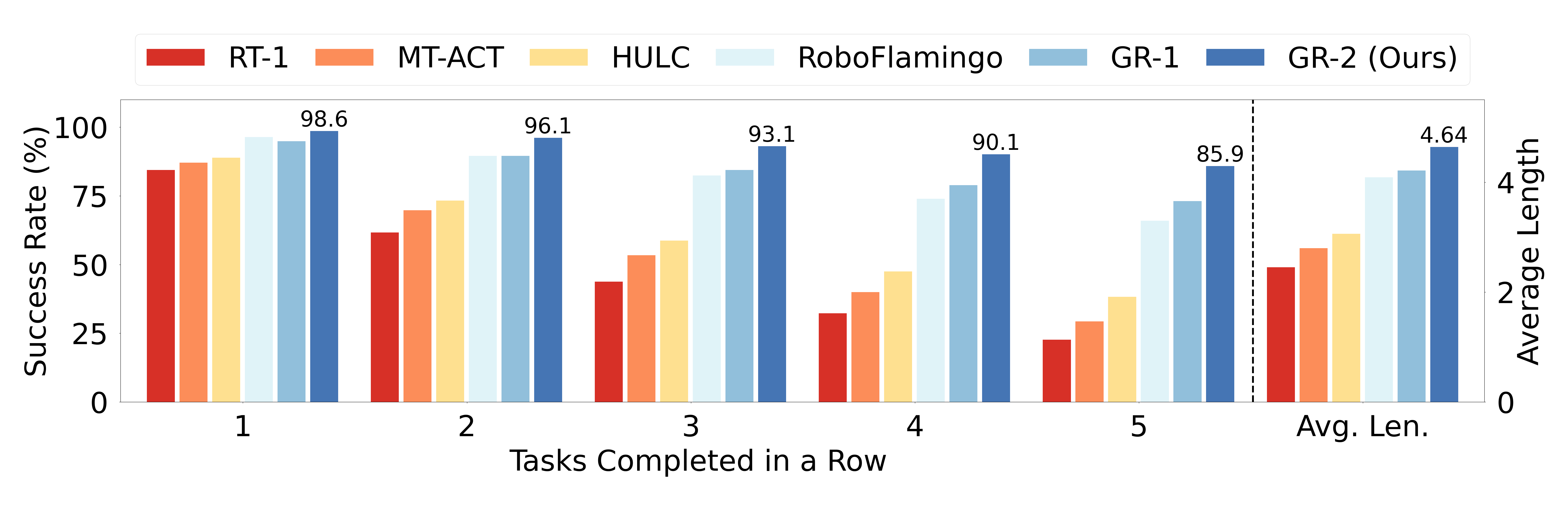}
    \caption{
        \textbf{CALVIN Benchmark Results.}
        We show the success rates of completing 1, 2, 3, 4, and 5 tasks in a row and the average length.
        The average length shows the average number of tasks the robot is able to accomplish when instructed to perform 5 tasks in a row.
    }
    \label{fig:calvin_success_rate}
\end{figure}

Fig.~\ref{fig:calvin_success_rate} shows the success rates of completing 1, 2, 3, 4, and 5 tasks in a row and the average length.
The average length is a comprehensive evaluation metric which shows the average number of tasks the robot is able to accomplish in a sequence across the 1,000 evaluated sequences.
We compare with five state-of-the-art baseline methods: RT-1~\cite{brohan2022rt}, MT-ACT~\cite{bharadhwaj2023roboagent}, HULC~\cite{mees2022matters}, RoboFlamingo~\cite{li2023vision}, and GR-1~\cite{wu2023unleashing}.
RT-1~\cite{brohan2022rt} is a language-conditioned multi-task policy that encodes the language condition via FiLM layers.
MT-ACT~\cite{bharadhwaj2023roboagent} similarly uses FiLM layers to inject the language condition and leverages an action-chunking transformer to address the multi-modality in the action data.
HULC~\cite{mees2022matters} is a hierarchical method which first predicts a plan in a latent space and uses the predicted plan for generating actions.
RoboFlamingo~\cite{li2023vision} fine-tunes a large pre-trained vision-language model on robotics data to perform language-conditioned manipulation.
\ourmethod establishes a new state of the art.
It outperforms all the comparing baseline methods in terms of success rates and the average length.
It improves the success rate of GR-1 from 94.9\% to 98.6\% for 1 task and from 73.1\% to 85.9\% for 5 tasks.
The average length is increased from 4.21 to 4.64.

\begin{figure}[t]
    \centering
    \includegraphics[width=\columnwidth]{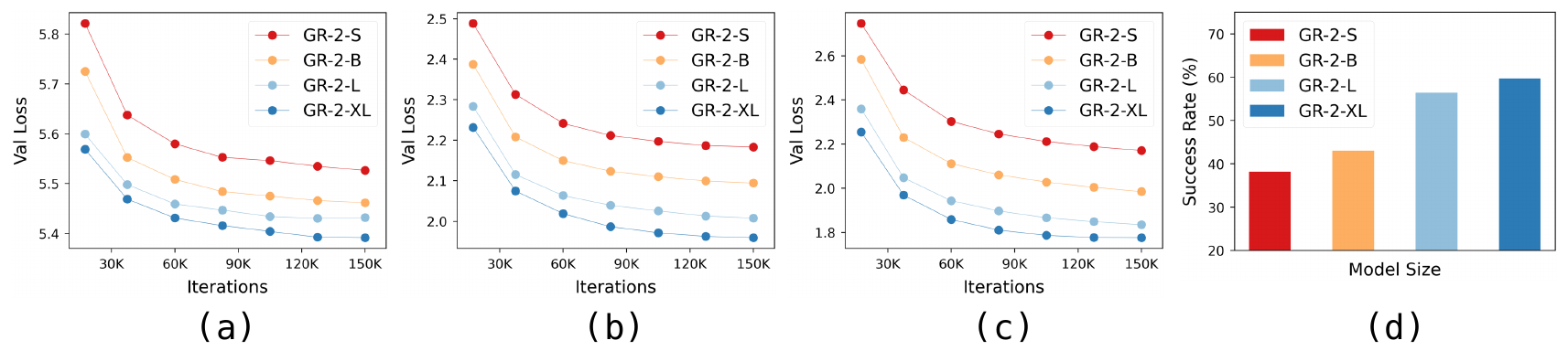}
    \caption{
        \textbf{Scaling Experiments.}
        We show the validation loss of video generation during pre-training on the validation sets of (a) Ego4D~\cite{grauman2022ego4d}, (b) RT-1~\cite{brohan2022rt}, and (c) our robot data. 
        (d) shows success rates in real-robot experiments.
        See Sec.~\ref{sec:exp:scaling} for more details.
    }
    \label{fig:scaling}
\end{figure}

\subsection{Autoregressive Video Generation}
\begin{figure}[htbp]
    \centering
    \includegraphics[width=1.\columnwidth, center, page=1]{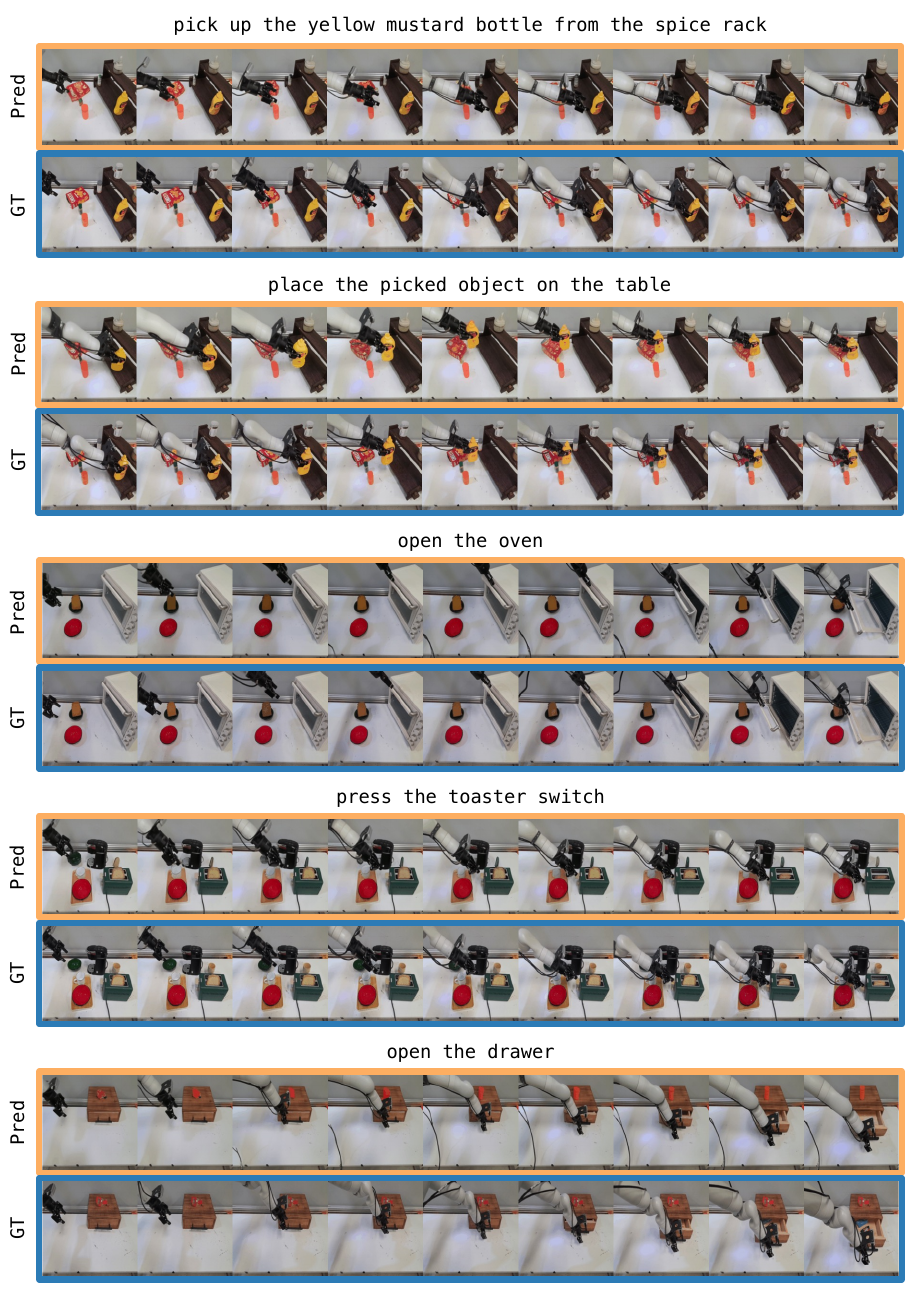}
    \caption{
        \textbf{Video Prediction (Pred) and Ground-Truth (GT) Rollouts of Multi-Task Learning (I).}
        We show autoregressive video predictions alongside the corresponding ground-truth videos captured from real-world rollouts.
    }
    \label{fig:multi_task_rollout_video_prediction1}
\end{figure}

\begin{figure}[htbp]
    \centering
    \includegraphics[width=\columnwidth, center, page=2]{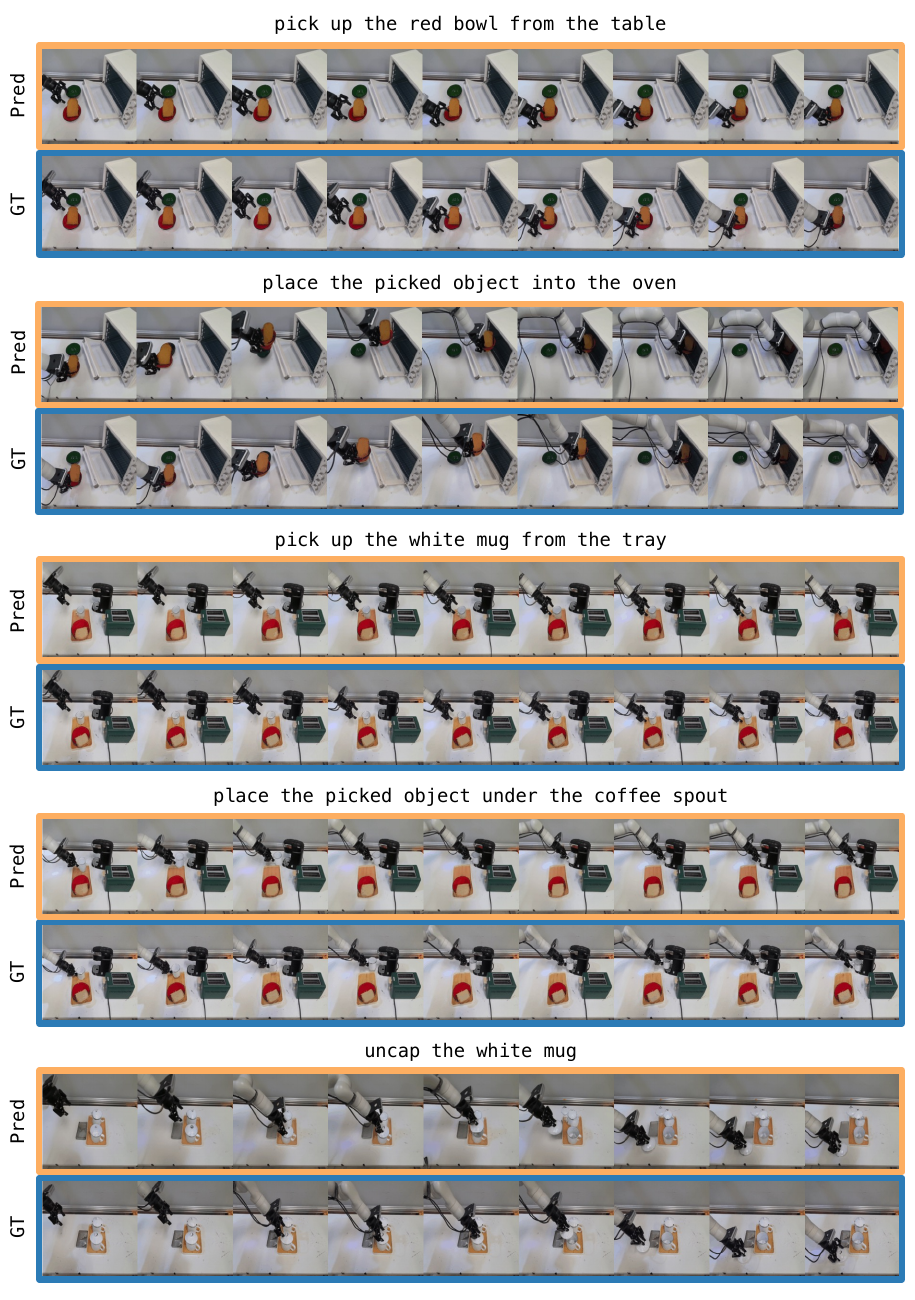}
    \caption{
        \textbf{Video Prediction (Pred) and Ground-Truth (GT) Rollouts of Multi-Task Learning (II).}
        We show autoregressive video predictions alongside the corresponding ground-truth videos captured from real-world rollouts.
    }
    \label{fig:multi_task_rollout_video_prediction2}
\end{figure}

\begin{figure}[htbp]
    \centering
    \includegraphics[width=0.95\columnwidth, center, page=1]{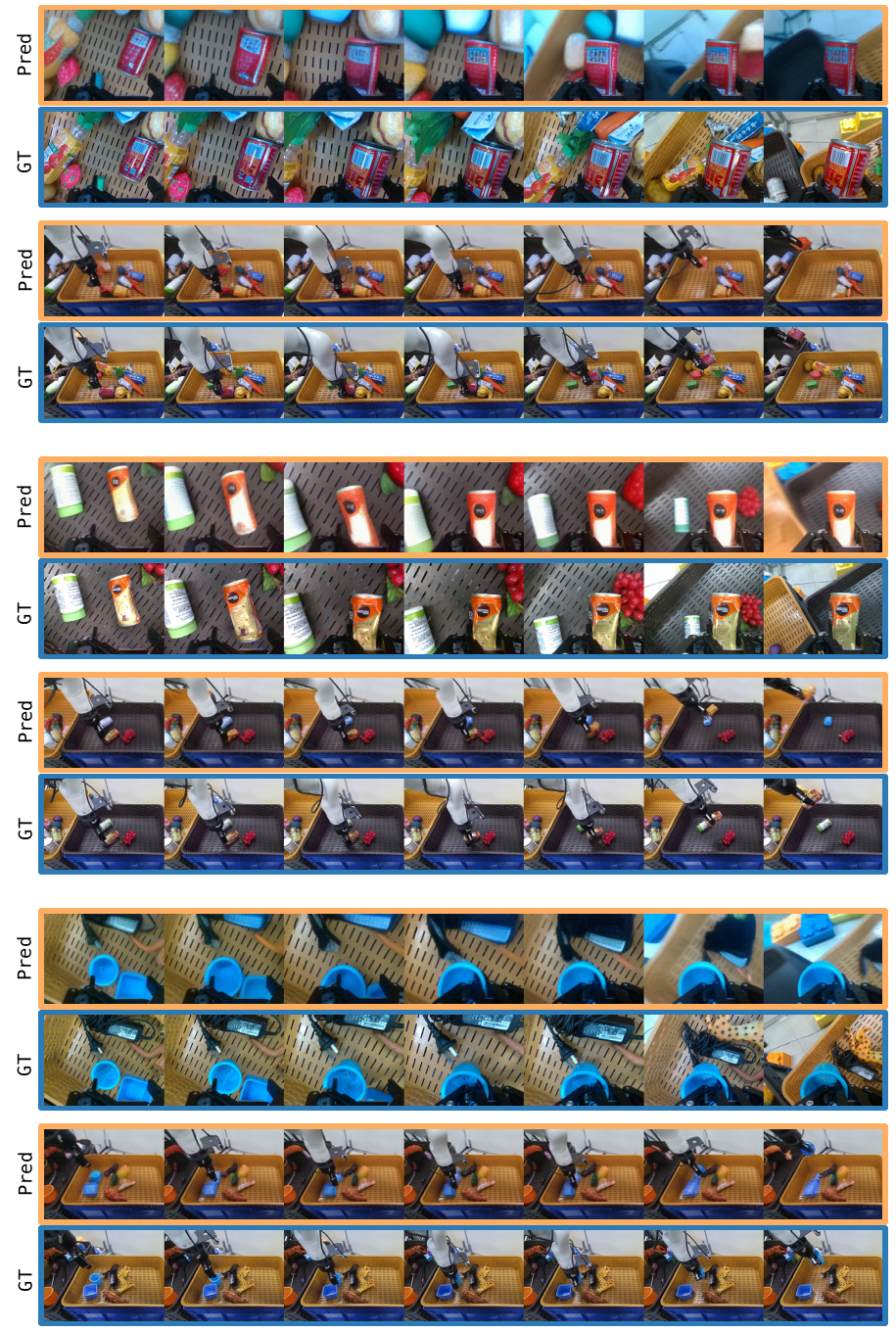}
    \caption{
        \textbf{Video Prediction (Pred) and Ground-Truth (GT) Rollouts of End-to-End Bin Picking (I).}
        We show autoregressive video predictions alongside the corresponding ground-truth videos captured from real-world rollouts.
        Both the views from the hand camera and the static head camera are shown.
    }
    \label{fig:bin_picking_rollout_video_prediction1}
\end{figure}

\begin{figure}[htbp]
    \centering
    \includegraphics[width=0.95\columnwidth, center, page=2]{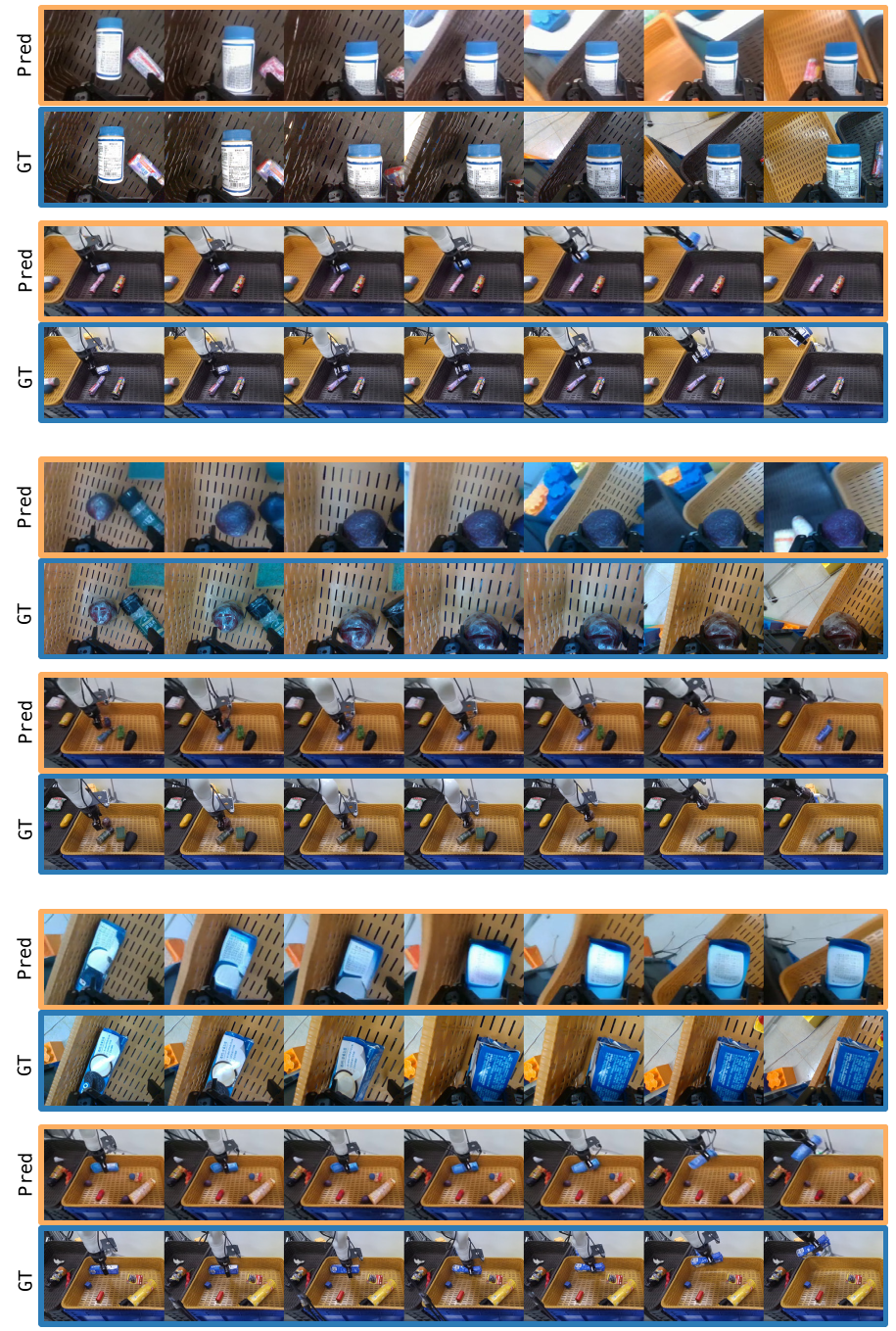}
    \caption{
        \textbf{Video Prediction (Pred) and Ground-Truth (GT) Rollouts of End-to-End Bin Picking (II).}
        We show autoregressive video predictions alongside the corresponding ground-truth videos captured from real-world rollouts.
        Both the views from the hand camera and the static head camera are shown.
    }
    \label{fig:bin_picking_rollout_video_prediction2}
\end{figure}

\begin{figure}[htbp]
\vspace{-1cm}
    \centering
    \includegraphics[width=0.95\columnwidth, center, page=1]{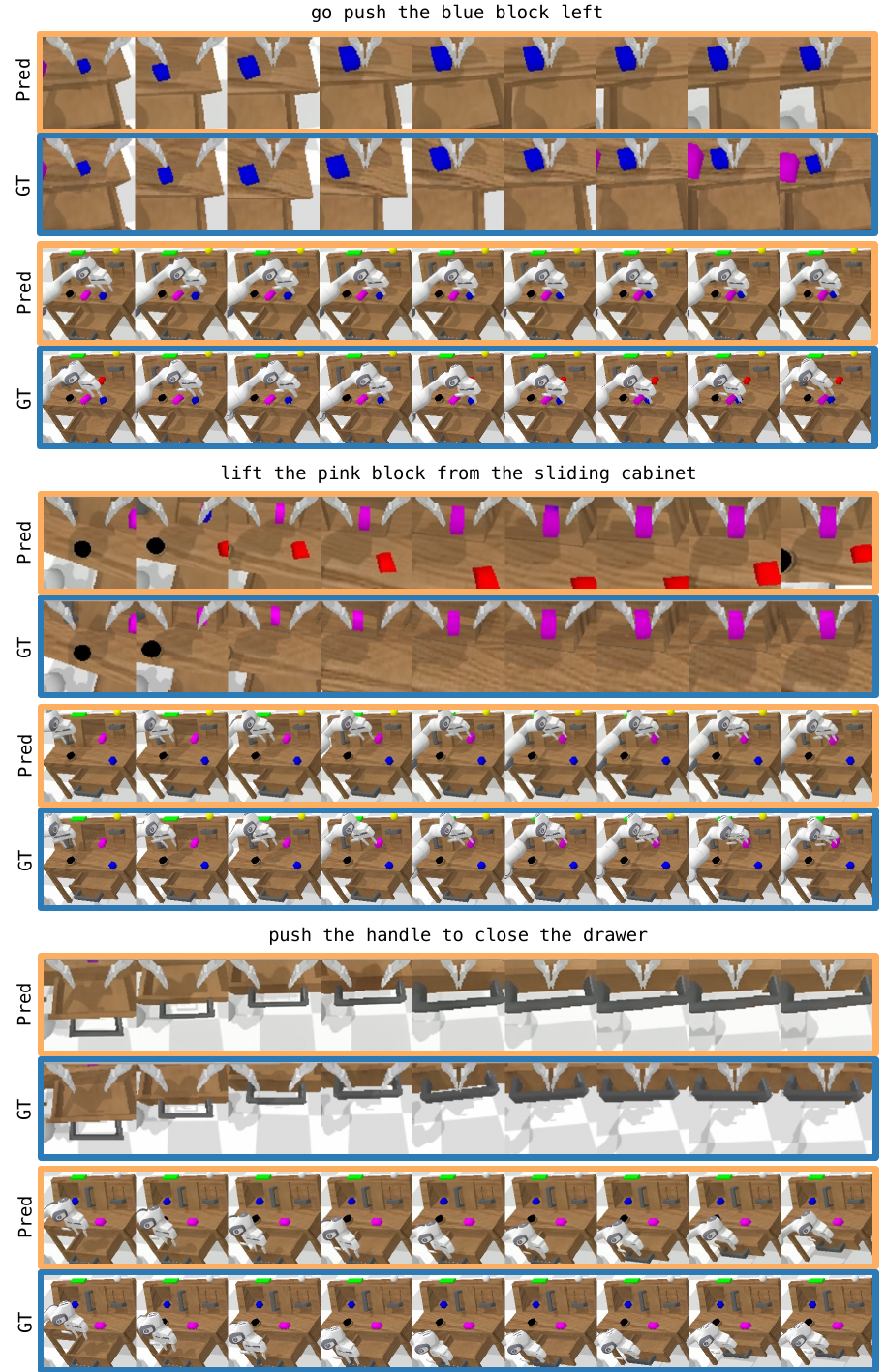}
    \caption{
        \textbf{Video Prediction (Pred) and Ground-Truth (GT) Rollouts of CALVIN Benchmark (I).}
        We show autoregressive video predictions alongside the corresponding ground-truth videos captured from the rollouts.
        Both the views from the hand camera and the static camera are shown.
    }
    \label{fig:calvin_rollout_video_prediction1}
\end{figure}

\begin{figure}[htbp]
\vspace{-1cm}
    \centering
    \includegraphics[width=0.95\columnwidth, center, page=2]{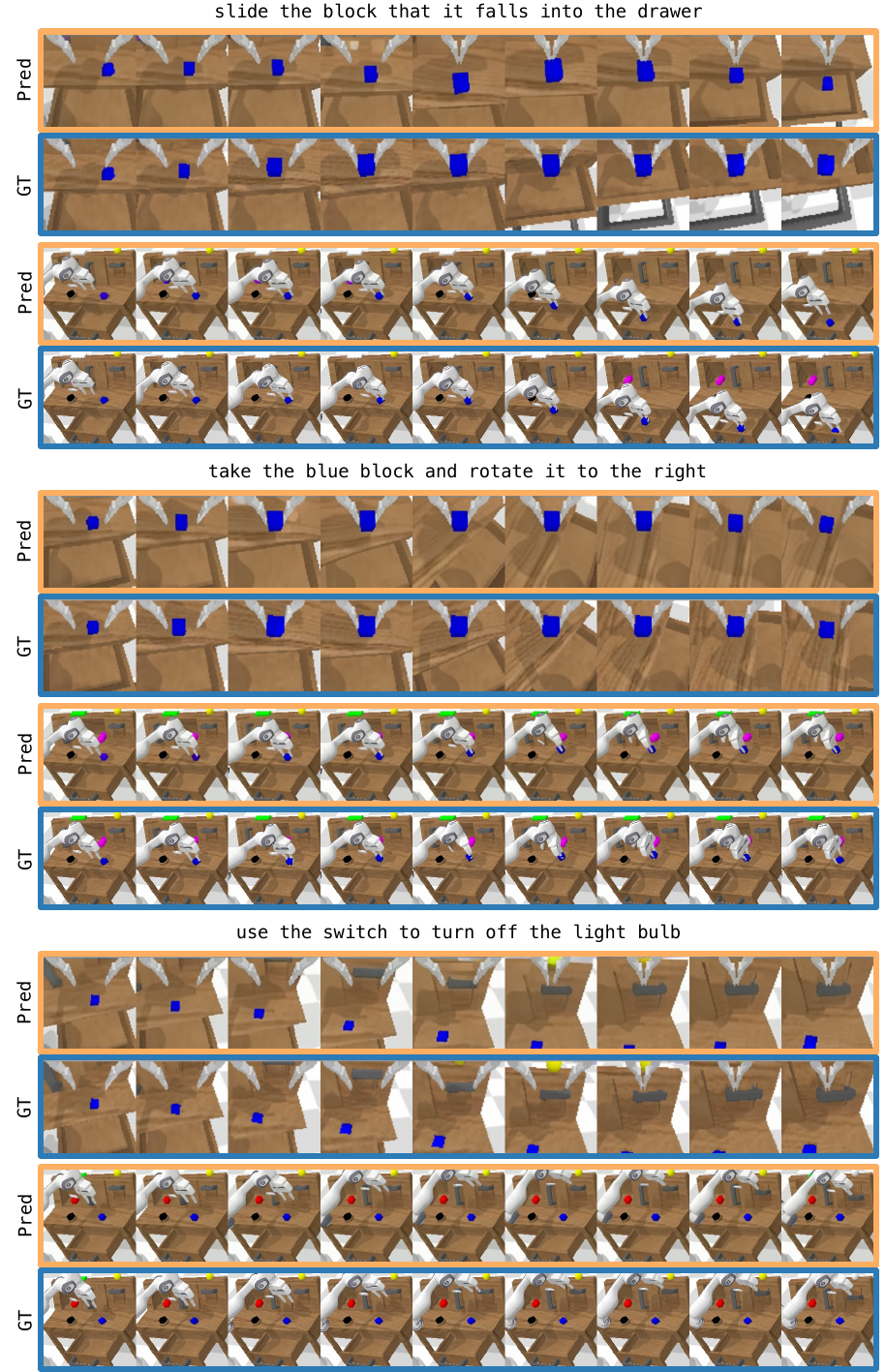}
    \caption{
        \textbf{Video Prediction (Pred) and Ground-Truth (GT) Rollouts of CALVIN Benchmark (II).}
        We show autoregressive video predictions alongside the corresponding ground-truth videos captured from the rollouts.
        Both the views from the hand camera and the static camera are shown.
    }
    \label{fig:calvin_rollout_video_prediction2}
\end{figure}

\ourmethod is pre-trained on a vast number of diverse videos, enabling it to predict future states within the image space. 
As a result, this video generation capability can effectively act as a planner for action generation.
That is, after generating the visual trajectory, an action trajectory can be subsequently inferred based on the visual trajectory.
To further investigate the effectiveness of this design, we visualize the video prediction result and compare it with the corresponding real rollout.
We show the visualization of rollouts from multi-task learning (Fig.~\ref{fig:multi_task_rollout_video_prediction1}~\ref{fig:multi_task_rollout_video_prediction2}), end-to-end bin picking (Fig.~\ref{fig:bin_picking_rollout_video_prediction1}~\ref{fig:bin_picking_rollout_video_prediction2}), and CALVIN (Fig.~\ref{fig:calvin_rollout_video_prediction1}~\ref{fig:calvin_rollout_video_prediction2}).

\ourmethod is able to generate high-quality videos alongside actions.
We highlight that the generated videos align with the real-world rollouts faithfully.
This indicates that the predicted action is trying to "replay" the trajectory in the predicted video.
This property brings about a simple approach to continuously improving action prediction by iteratively improving video generation.

\subsection{Scaling}
\label{sec:exp:scaling}
We investigate how scaling up the model size can help \ourmethod in pre-training and fine-tuning.
In particular, we pre-train \ourmethod of four sizes.
The number of trainable parameters is 30M (\ourmethod-S), 95M (\ourmethod-B), 312M (\ourmethod-L), and 719M (\ourmethod-XL), respectively.
The validation loss of video prediction is shown in Fig.~\ref{fig:scaling}(a)(b)(c).
The validation loss decreases with the increase of the model size, showing scalability in terms of video generation.
We incorporate videos of in-domain robot data during the pre-training stage and keep the pre-trained parameters frozen while fine-tuning on robot trajectories.
After fine-tuning, we evaluate different models on a subset of settings in Sec.~\ref{sec:exp:multi-task}.
Results are shown in Fig.~\ref{fig:scaling}(d).
The success rate scales well with the model size.
This result highlights the strong potential of \ourmethod for continuous performance improvement through increasing the model size.

\section{Related Work}
\label{sec:related_work}
\textbf{Generalist Robot Manipulation.}
A long-standing goal in robotics research is to develop a generalist robot agent that is able to accomplish a wide range of tasks in diverse environments.
One of the most flexible ways to specify tasks is through natural languages~\cite{brohan2022rt, brohan2023rt,wu2023unleashing, li2023vision, jiang2022vima, lynch2020language, mees2022matters, bharadhwaj2023roboagent, jang2022bc, reuss2024multimodal, ha2023scaling, shridhar2022cliport, team2024octo, black2023zero}.
Pioneering studies explored using large-scale robot datasets to learn generalist policies which are able to accomplish a variety of tasks~\cite{jang2022bc, brohan2022rt, brohan2023rt, padalkar2023open}.
To achieve generalization in unseen scenarios, some existing works combined data from other domains with robot data in policy training~\cite{reed2022generalist, wu2023unleashing, brohan2023rt}.
Recently, a number of works proposed to fine-tune a vision-language model which has been pre-trained on Internet-scaled data to obtain robust and generalizable robot policies~\cite{li2023vision, brohan2023rt, kim2024openvla}.
In addition, some recent works resort to 3D information~\cite{shridhar2023perceiver, xian2023chaineddiffuser, ke20243d, gervet2023act3d} to achieve efficient policy learning by leveraging the geometry information contained in 3D data.
Another line of works proposed to condition the policy with a goal image instead of a language~\cite{bousmalis2023robocat, wu2022transporters, seita2021learning, groth2021goal, davchev2021wish}.
And previous methods have also explored aligning the latent space of goal images and languages to enable both goal image condition and language condition during training~\cite{mees2022matters, lynch2020language, reuss2024multimodal}.
\ourmethod is a language-conditioned generalist robot manipulation agent.
Unlike most previous works, it is first pre-trained on video generation with Internet-scale video datasets and then fine-tuned on robot data to predict both actions and videos.

\textbf{Robot Learning with Pre-training.}
Inspired by the success in the field of vision~\cite{he2022masked} and languages~\cite{brown2020languagemodelsfewshotlearners}, pre-training has gained increasing popularity in robot learning as it enhances the generalization capabilities and robustness of policies~\cite{xiao2022masked, karamcheti2023language, nair2022r3m, radosavovic2023robot, hafner2023mastering, wen2023any, yang2023learning, kim2024openvla, wu2023unleashing}.
A popular approach is to first learn useful visual representations via masked modeling~\cite{karamcheti2023language, seo2023masked, radosavovic2023real, xiao2022masked} or contrastive learning~\cite{nair2022r3m, jing2023exploring, laskin2020curl, sermanet2018time}.
The learned representations are then used for downstream policy learning.
RPT~\cite{radosavovic2023robot} performed self-supervised pre-training and showcased that pre-training with large robot datasets consistently surpasses training from scratch.
In reinforcement learning (RL), previous works proposed to first train a world model to obtain latent state representations and then use them to train an RL agent~\cite{ha2018world, hafner2023mastering, seo2023masked}.
VIPER~\cite{escontrela2024video} trained a video prediction model with expert data and utilized it as an action-free reward signal to train RL policies.
Some model-based methods trained a video prediction model and combined it with an inverse dynamics model~\cite{du2024learning, yang2023learning, du2023video} or model predictive control~\cite{finn2017deep, gupta2022maskvit} to perform robot manipulation.
VPT~\cite{baker2022video} first trained an inverse dynamics model with a small amount of data labeled with actions and used it to label a large amount of unlabeled data gathered from the web for policy training in Minecraft.
Recent works trained policies based on models that have been pre-trained on Internet-scale data via end-to-end fine-tuning~\cite{li2023vision}, co-training with robot data~\cite{brohan2023rt, kim2024openvla}, or a two-stream architecture~\cite{lin2024spawnnet}.
The policy can make use of the web-scale knowledge obtained from pre-training in policy learning and showcases powerful generalization capabilities in unseen scenarios.
Inspired by these works, we propose to leverage large-scale text-video data to perform video generative pre-training in our previous work GR-1~\cite{wu2023unleashing}.
The motivation is that we believe videos contain valuable information on the dynamics of the environment and how the environment should evolve according to the text description.
This information can facilitate action prediction during downstream policy learning.
In comparison to GR-1~\cite{wu2023unleashing}, \ourmethod scales the number of pre-training videos from 0.8 million to 38 million, boosting the generalization capabilities in various unseen scenarios.
In addition, the innovative model architecture facilitates more seamless knowledge transfer between pre-training and fine-tuning, leading to a policy that is more generalizable and robust.

\section{Conclusions}
\label{sec:conclusion}
We present \ourmethod, a generative robotic video-language-action model that is able to effectively learn a wide variety of tasks and generalize to unseen scenarios.
\ourmethod is first pre-trained on video generation with 38 million Internet videos.
It is then fine-tuned on robot data to predict action trajectories and videos in tandem.
It showcases strong multi-task learning capabilities, successfully completing more than 100 different manipulation tasks in the real world with a high success rate.
It generalizes well to novel scenarios, including unseen backgrounds, environments, objects, and tasks.
Moreover, \ourmethod can perform bin-picking manipulation with over 100 objects in an end-to-end manner and handle unseen objects with remarkable robustness.
We observe a strong correlation between the generated video and the action predicted alongside.
In the future, we plan to enhance the generalization capabilities and robustness of action prediction, with a particular focus on improving the performance on unseen manipulation.

\section*{Contributions \& Acknowledgements}
\label{sec:author}
\begin{enumerate}[leftmargin=*]
    \item[$\bullet$] \textbf{Evaluation}: Chi-Lam Cheang, Yuxiao Liu, Hongtao Wu, Jiafeng Xu, Yichu Yang, Minzhao Zhu
    \item[$\bullet$] \textbf{Model \& Training}: Ya Jing, Tao Kong, Yuxiao Liu, Hongtao Wu, Yichu Yang, Hanbo Zhang, Minzhao Zhu
    \item[$\bullet$] \textbf{Data Collection \& Curation}: Chi-Lam Cheang, Guangzeng Chen, Ya Jing, Tao Kong, Yifeng Li, Hongtao Wu, Jiafeng Xu, Minzhao Zhu
    \item[$\bullet$] \textbf{Paper Writing}: Ya Jing, Tao Kong, Hang Li, Hongtao Wu, Yichu Yang
\end{enumerate}

We thank the engineering team at ByteDance for their outstanding technical skills, and the data team for their meticulous work on data collection, annotation, and processing. 
We thank Xiao Ma for his helpful discussion and valuable advice on paper writing. 
Their commitment and expertise are crucial to the success of this project.

\bibliography{reference}

\begin{thebibliography}{10}

\bibitem{achiam2023gpt}
Josh Achiam, Steven Adler, Sandhini Agarwal, Lama Ahmad, Ilge Akkaya, Florencia~Leoni Aleman, Diogo Almeida, Janko Altenschmidt, Sam Altman, Shyamal Anadkat, et~al.
\newblock {GPT}-4 technical report.
\newblock {\em arXiv preprint arXiv:2303.08774}, 2023.

\bibitem{ravi2024sam}
Nikhila Ravi, Valentin Gabeur, Yuan-Ting Hu, Ronghang Hu, Chaitanya Ryali, Tengyu Ma, Haitham Khedr, Roman R{\"a}dle, Chloe Rolland, Laura Gustafson, et~al.
\newblock {SAM 2}: Segment anything in images and videos.
\newblock {\em arXiv preprint arXiv:2408.00714}, 2024.

\bibitem{videoworldsimulators2024}
Tim Brooks, Bill Peebles, Connor Holmes, Will DePue, Yufei Guo, Li~Jing, David Schnurr, Joe Taylor, Troy Luhman, Eric Luhman, Clarence Ng, Ricky Wang, and Aditya Ramesh.
\newblock Video generation models as world simulators.
\newblock 2024.

\bibitem{brown2020language}
Tom~B. Brown, Benjamin Mann, Nick Ryder, Melanie Subbiah, Jared Kaplan, Prafulla Dhariwal, Arvind Neelakantan, Pranav Shyam, Girish Sastry, Amanda Askell, et~al.
\newblock Language models are few-shot learners.
\newblock {\em arXiv preprint arXiv:2005.14165}, 2020.

\bibitem{wu2023unleashing}
Hongtao Wu, Ya~Jing, Chilam Cheang, Guangzeng Chen, Jiafeng Xu, Xinghang Li, Minghuan Liu, Hang Li, and Tao Kong.
\newblock Unleashing large-scale video generative pre-training for visual robot manipulation.
\newblock {\em arXiv preprint arXiv:2312.13139}, 2023.

\bibitem{radford2021learning}
Alec Radford, Jong~Wook Kim, Chris Hallacy, Aditya Ramesh, Gabriel Goh, Sandhini Agarwal, Girish Sastry, Amanda Askell, Pamela Mishkin, Jack Clark, et~al.
\newblock Learning transferable visual models from natural language supervision.
\newblock In {\em International conference on machine learning}, pages 8748--8763. PMLR, 2021.

\bibitem{esser2021taming}
Patrick Esser, Robin Rombach, and Bjorn Ommer.
\newblock Taming transformers for high-resolution image synthesis.
\newblock In {\em Proceedings of the IEEE/CVF conference on computer vision and pattern recognition}, pages 12873--12883, 2021.

\bibitem{miech2019howto100m}
Antoine Miech, Dimitri Zhukov, Jean-Baptiste Alayrac, Makarand Tapaswi, Ivan Laptev, and Josef Sivic.
\newblock Howto100m: Learning a text-video embedding by watching hundred million narrated video clips.
\newblock In {\em Proceedings of the IEEE/CVF international conference on computer vision}, pages 2630--2640, 2019.

\bibitem{grauman2022ego4d}
Kristen Grauman, Andrew Westbury, Eugene Byrne, Zachary Chavis, Antonino Furnari, Rohit Girdhar, Jackson Hamburger, Hao Jiang, Miao Liu, Xingyu Liu, et~al.
\newblock {Ego4D}: Around the world in 3,000 hours of egocentric video.
\newblock In {\em Proceedings of the IEEE/CVF Conference on Computer Vision and Pattern Recognition}, pages 18995--19012, 2022.

\bibitem{goyal2017something}
Raghav Goyal, Samira Ebrahimi~Kahou, Vincent Michalski, Joanna Materzynska, Susanne Westphal, Heuna Kim, Valentin Haenel, Ingo Fruend, Peter Yianilos, Moritz Mueller-Freitag, et~al.
\newblock The" something something" video database for learning and evaluating visual common sense.
\newblock In {\em Proceedings of the IEEE international conference on computer vision}, pages 5842--5850, 2017.

\bibitem{damen2018scaling}
Dima Damen, Hazel Doughty, Giovanni~Maria Farinella, Sanja Fidler, Antonino Furnari, Evangelos Kazakos, Davide Moltisanti, Jonathan Munro, Toby Perrett, Will Price, et~al.
\newblock Scaling egocentric vision: The epic-kitchens dataset.
\newblock In {\em Proceedings of the European conference on computer vision (ECCV)}, pages 720--736, 2018.

\bibitem{carreira2019short}
Joao Carreira, Eric Noland, Chloe Hillier, and Andrew Zisserman.
\newblock A short note on the kinetics-700 human action dataset.
\newblock {\em arXiv preprint arXiv:1907.06987}, 2019.

\bibitem{lugaresi2019mediapipe}
Camillo Lugaresi, Jiuqiang Tang, Hadon Nash, Chris McClanahan, Esha Uboweja, Michael Hays, Fan Zhang, Chuo-Ling Chang, Ming~Guang Yong, Juhyun Lee, et~al.
\newblock Mediapipe: A framework for building perception pipelines.
\newblock {\em arXiv preprint arXiv:1906.08172}, 2019.

\bibitem{opensora}
Zangwei Zheng, Xiangyu Peng, Tianji Yang, Chenhui Shen, Shenggui Li, Hongxin Liu, Yukun Zhou, Tianyi Li, and Yang You.
\newblock {Open-Sora}: Democratizing efficient video production for all, March 2024.

\bibitem{brohan2022rt}
Anthony Brohan, Noah Brown, Justice Carbajal, Yevgen Chebotar, Joseph Dabis, Chelsea Finn, Keerthana Gopalakrishnan, Karol Hausman, Alex Herzog, Jasmine Hsu, et~al.
\newblock {RT-1}: Robotics transformer for real-world control at scale.
\newblock {\em arXiv preprint arXiv:2212.06817}, 2022.

\bibitem{walke2023bridgedata}
Homer~Rich Walke, Kevin Black, Tony~Z Zhao, Quan Vuong, Chongyi Zheng, Philippe Hansen-Estruch, Andre~Wang He, Vivek Myers, Moo~Jin Kim, Max Du, et~al.
\newblock Bridgedata v2: A dataset for robot learning at scale.
\newblock In {\em Conference on Robot Learning}, pages 1723--1736. PMLR, 2023.

\bibitem{sohn2015learning}
Kihyuk Sohn, Honglak Lee, and Xinchen Yan.
\newblock Learning structured output representation using deep conditional generative models.
\newblock {\em Advances in neural information processing systems}, 28, 2015.

\bibitem{kingma2013auto}
Diederik~P Kingma and Max Welling.
\newblock Auto-encoding variational bayes.
\newblock {\em arXiv preprint arXiv:1312.6114}, 2013.

\bibitem{zhao2023learning}
Tony~Z Zhao, Vikash Kumar, Sergey Levine, and Chelsea Finn.
\newblock Learning fine-grained bimanual manipulation with low-cost hardware.
\newblock {\em arXiv preprint arXiv:2304.13705}, 2023.

\bibitem{yang2023moma-force}
Taozheng Yang, Ya~Jing, Hongtao Wu, Jiafeng Xu, Kuankuan Sima, Guangzeng Chen, Qie Sima, and Tao Kong.
\newblock {MOMA-Force}: Visual-force imitation for real-world mobile manipulation.
\newblock In {\em 2023 IEEE/RSJ International Conference on Intelligent Robots and Systems (IROS)}, pages 6847--6852. IEEE, 2023.

\bibitem{mees2022calvin}
Oier Mees, Lukas Hermann, Erick Rosete-Beas, and Wolfram Burgard.
\newblock {CALVIN}: A benchmark for language-conditioned policy learning for long-horizon robot manipulation tasks.
\newblock {\em IEEE Robotics and Automation Letters (RA-L)}, 7(3):7327--7334, 2022.

\bibitem{ho2020denoising}
Jonathan Ho, Ajay Jain, and Pieter Abbeel.
\newblock Denoising diffusion probabilistic models.
\newblock {\em Advances in neural information processing systems}, 33:6840--6851, 2020.

\bibitem{kuznetsova2020open}
Alina Kuznetsova, Hassan Rom, Neil Alldrin, Jasper Uijlings, Ivan Krasin, Jordi Pont-Tuset, Shahab Kamali, Stefan Popov, Matteo Malloci, Alexander Kolesnikov, et~al.
\newblock The open images dataset v4: Unified image classification, object detection, and visual relationship detection at scale.
\newblock {\em International journal of computer vision}, 128(7):1956--1981, 2020.

\bibitem{kirillov2023segment}
Alexander Kirillov, Eric Mintun, Nikhila Ravi, Hanzi Mao, Chloe Rolland, Laura Gustafson, Tete Xiao, Spencer Whitehead, Alexander~C Berg, Wan-Yen Lo, et~al.
\newblock {Segment anything}.
\newblock In {\em Proceedings of the IEEE/CVF International Conference on Computer Vision}, pages 4015--4026, 2023.

\bibitem{ma2024latte}
Xin Ma, Yaohui Wang, Gengyun Jia, Xinyuan Chen, Ziwei Liu, Yuan-Fang Li, Cunjian Chen, and Yu~Qiao.
\newblock Latte: Latent diffusion transformer for video generation.
\newblock {\em arXiv preprint arXiv:2401.03048}, 2024.

\bibitem{bharadhwaj2023roboagent}
Homanga Bharadhwaj, Jay Vakil, Mohit Sharma, Abhinav Gupta, Shubham Tulsiani, and Vikash Kumar.
\newblock {RoboAgent}: Generalization and efficiency in robot manipulation via semantic augmentations and action chunking.
\newblock {\em arXiv preprint arXiv:2309.01918}, 2023.

\bibitem{mees2022matters}
Oier Mees, Lukas Hermann, and Wolfram Burgard.
\newblock What matters in language conditioned robotic imitation learning over unstructured data.
\newblock {\em IEEE Robotics and Automation Letters}, 7(4):11205--11212, 2022.

\bibitem{li2023vision}
Xinghang Li, Minghuan Liu, Hanbo Zhang, Cunjun Yu, Jie Xu, Hongtao Wu, Chilam Cheang, Ya~Jing, Weinan Zhang, Huaping Liu, et~al.
\newblock Vision-language foundation models as effective robot imitators.
\newblock {\em arXiv preprint arXiv:2311.01378}, 2023.

\bibitem{brohan2023rt}
Anthony Brohan, Noah Brown, Justice Carbajal, Yevgen Chebotar, Xi~Chen, Krzysztof Choromanski, Tianli Ding, Danny Driess, Avinava Dubey, Chelsea Finn, et~al.
\newblock {RT-2}: Vision-language-action models transfer web knowledge to robotic control.
\newblock {\em arXiv preprint arXiv:2307.15818}, 2023.

\bibitem{jiang2022vima}
Yunfan Jiang, Agrim Gupta, Zichen Zhang, Guanzhi Wang, Yongqiang Dou, Yanjun Chen, Li~Fei-Fei, Anima Anandkumar, Yuke Zhu, and Linxi Fan.
\newblock {VIMA}: General robot manipulation with multimodal prompts.
\newblock {\em arXiv preprint arXiv:2210.03094}, 2(3):6, 2022.

\bibitem{lynch2020language}
Corey Lynch and Pierre Sermanet.
\newblock Language conditioned imitation learning over unstructured data.
\newblock {\em arXiv preprint arXiv:2005.07648}, 2020.

\bibitem{jang2022bc}
Eric Jang, Alex Irpan, Mohi Khansari, Daniel Kappler, Frederik Ebert, Corey Lynch, Sergey Levine, and Chelsea Finn.
\newblock {BC-Z}: Zero-shot task generalization with robotic imitation learning.
\newblock In {\em Conference on Robot Learning}, pages 991--1002. PMLR, 2022.

\bibitem{reuss2024multimodal}
Moritz Reuss, {\"O}mer~Erdin{\c{c}} Ya{\u{g}}murlu, Fabian Wenzel, and Rudolf Lioutikov.
\newblock Multimodal diffusion transformer: Learning versatile behavior from multimodal goals.
\newblock In {\em Robotics: Science and Systems}, 2024.

\bibitem{ha2023scaling}
Huy Ha, Pete Florence, and Shuran Song.
\newblock Scaling up and distilling down: Language-guided robot skill acquisition.
\newblock In {\em Conference on Robot Learning}, pages 3766--3777. PMLR, 2023.

\bibitem{shridhar2022cliport}
Mohit Shridhar, Lucas Manuelli, and Dieter Fox.
\newblock {CLIPort}: What and where pathways for robotic manipulation.
\newblock In {\em Conference on robot learning}, pages 894--906. PMLR, 2022.

\bibitem{team2024octo}
Octo~Model Team, Dibya Ghosh, Homer Walke, Karl Pertsch, Kevin Black, Oier Mees, Sudeep Dasari, Joey Hejna, Tobias Kreiman, Charles Xu, et~al.
\newblock Octo: An open-source generalist robot policy.
\newblock {\em arXiv preprint arXiv:2405.12213}, 2024.

\bibitem{black2023zero}
Kevin Black, Mitsuhiko Nakamoto, Pranav Atreya, Homer Walke, Chelsea Finn, Aviral Kumar, and Sergey Levine.
\newblock Zero-shot robotic manipulation with pretrained image-editing diffusion models.
\newblock {\em arXiv preprint arXiv:2310.10639}, 2023.

\bibitem{padalkar2023open}
Abhishek Padalkar, Acorn Pooley, Ajinkya Jain, Alex Bewley, Alex Herzog, Alex Irpan, Alexander Khazatsky, Anant Rai, Anikait Singh, Anthony Brohan, et~al.
\newblock Open x-embodiment: Robotic learning datasets and rt-x models.
\newblock {\em arXiv preprint arXiv:2310.08864}, 2023.

\bibitem{reed2022generalist}
Scott Reed, Konrad Zolna, Emilio Parisotto, Sergio~Gomez Colmenarejo, Alexander Novikov, Gabriel Barth-Maron, Mai Gimenez, Yury Sulsky, Jackie Kay, Jost~Tobias Springenberg, et~al.
\newblock A generalist agent.
\newblock {\em arXiv preprint arXiv:2205.06175}, 2022.

\bibitem{kim2024openvla}
Moo~Jin Kim, Karl Pertsch, Siddharth Karamcheti, Ted Xiao, Ashwin Balakrishna, Suraj Nair, Rafael Rafailov, Ethan Foster, Grace Lam, Pannag Sanketi, et~al.
\newblock {OpenVLA}: An open-source vision-language-action model.
\newblock {\em arXiv preprint arXiv:2406.09246}, 2024.

\bibitem{shridhar2023perceiver}
Mohit Shridhar, Lucas Manuelli, and Dieter Fox.
\newblock Perceiver-actor: A multi-task transformer for robotic manipulation.
\newblock In {\em Conference on Robot Learning}, pages 785--799. PMLR, 2023.

\bibitem{xian2023chaineddiffuser}
Zhou Xian, Nikolaos Gkanatsios, Theophile Gervet, Tsung-Wei Ke, and Katerina Fragkiadaki.
\newblock Chaineddiffuser: Unifying trajectory diffusion and keypose prediction for robotic manipulation.
\newblock In {\em 7th Annual Conference on Robot Learning}, 2023.

\bibitem{ke20243d}
Tsung-Wei Ke, Nikolaos Gkanatsios, and Katerina Fragkiadaki.
\newblock 3d diffuser actor: Policy diffusion with 3d scene representations.
\newblock {\em arXiv preprint arXiv:2402.10885}, 2024.

\bibitem{gervet2023act3d}
Theophile Gervet, Zhou Xian, Nikolaos Gkanatsios, and Katerina Fragkiadaki.
\newblock {Act3D}: 3d feature field transformers for multi-task robotic manipulation.
\newblock In {\em 7th Annual Conference on Robot Learning}, 2023.

\bibitem{bousmalis2023robocat}
Konstantinos Bousmalis, Giulia Vezzani, Dushyant Rao, Coline Devin, Alex~X Lee, Maria Bauza, Todor Davchev, Yuxiang Zhou, Agrim Gupta, Akhil Raju, et~al.
\newblock {RoboCat}: A self-improving foundation agent for robotic manipulation.
\newblock {\em arXiv preprint arXiv:2306.11706}, 2023.

\bibitem{wu2022transporters}
Hongtao Wu, Jikai Ye, Xin Meng, Chris Paxton, and Gregory~S Chirikjian.
\newblock Transporters with visual foresight for solving unseen rearrangement tasks.
\newblock In {\em 2022 IEEE/RSJ International Conference on Intelligent Robots and Systems (IROS)}, pages 10756--10763. IEEE, 2022.

\bibitem{seita2021learning}
Daniel Seita, Pete Florence, Jonathan Tompson, Erwin Coumans, Vikas Sindhwani, Ken Goldberg, and Andy Zeng.
\newblock Learning to rearrange deformable cables, fabrics, and bags with goal-conditioned transporter networks.
\newblock In {\em 2021 IEEE International Conference on Robotics and Automation (ICRA)}, pages 4568--4575. IEEE, 2021.

\bibitem{groth2021goal}
Oliver Groth, Chia-Man Hung, Andrea Vedaldi, and Ingmar Posner.
\newblock Goal-conditioned end-to-end visuomotor control for versatile skill primitives.
\newblock In {\em 2021 IEEE International Conference on Robotics and Automation (ICRA)}, pages 1319--1325. IEEE, 2021.

\bibitem{davchev2021wish}
Todor Davchev, Oleg Sushkov, Jean-Baptiste Regli, Stefan Schaal, Yusuf Aytar, Markus Wulfmeier, and Jon Scholz.
\newblock Wish you were here: Hindsight goal selection for long-horizon dexterous manipulation.
\newblock {\em arXiv preprint arXiv:2112.00597}, 2021.

\bibitem{he2022masked}
Kaiming He, Xinlei Chen, Saining Xie, Yanghao Li, Piotr Doll{\'a}r, and Ross Girshick.
\newblock Masked autoencoders are scalable vision learners.
\newblock In {\em Proceedings of the IEEE/CVF conference on computer vision and pattern recognition}, pages 16000--16009, 2022.

\bibitem{brown2020languagemodelsfewshotlearners}
Tom~B. Brown, Benjamin Mann, Nick Ryder, Melanie Subbiah, Jared Kaplan, Prafulla Dhariwal, Arvind Neelakantan, Pranav Shyam, Girish Sastry, Amanda Askell, et~al.
\newblock Language models are few-shot learners, 2020.

\bibitem{xiao2022masked}
Tete Xiao, Ilija Radosavovic, Trevor Darrell, and Jitendra Malik.
\newblock Masked visual pre-training for motor control.
\newblock {\em arXiv preprint arXiv:2203.06173}, 2022.

\bibitem{karamcheti2023language}
Siddharth Karamcheti, Suraj Nair, Annie~S Chen, Thomas Kollar, Chelsea Finn, Dorsa Sadigh, and Percy Liang.
\newblock Language-driven representation learning for robotics.
\newblock {\em arXiv preprint arXiv:2302.12766}, 2023.

\bibitem{nair2022r3m}
Suraj Nair, Aravind Rajeswaran, Vikash Kumar, Chelsea Finn, and Abhinav Gupta.
\newblock {R3M}: A universal visual representation for robot manipulation.
\newblock {\em arXiv preprint arXiv:2203.12601}, 2022.

\bibitem{radosavovic2023robot}
Ilija Radosavovic, Baifeng Shi, Letian Fu, Ken Goldberg, Trevor Darrell, and Jitendra Malik.
\newblock Robot learning with sensorimotor pre-training.
\newblock In {\em Conference on Robot Learning}, pages 683--693. PMLR, 2023.

\bibitem{hafner2023mastering}
Danijar Hafner, Jurgis Pasukonis, Jimmy Ba, and Timothy Lillicrap.
\newblock Mastering diverse domains through world models.
\newblock {\em arXiv preprint arXiv:2301.04104}, 2023.

\bibitem{wen2023any}
Chuan Wen, Xingyu Lin, John So, Kai Chen, Qi~Dou, Yang Gao, and Pieter Abbeel.
\newblock Any-point trajectory modeling for policy learning.
\newblock {\em arXiv preprint arXiv:2401.00025}, 2023.

\bibitem{yang2023learning}
Mengjiao Yang, Yilun Du, Kamyar Ghasemipour, Jonathan Tompson, Dale Schuurmans, and Pieter Abbeel.
\newblock Learning interactive real-world simulators.
\newblock {\em arXiv preprint arXiv:2310.06114}, 2023.

\bibitem{seo2023masked}
Younggyo Seo, Danijar Hafner, Hao Liu, Fangchen Liu, Stephen James, Kimin Lee, and Pieter Abbeel.
\newblock Masked world models for visual control.
\newblock In {\em Conference on Robot Learning}, pages 1332--1344. PMLR, 2023.

\bibitem{radosavovic2023real}
Ilija Radosavovic, Tete Xiao, Stephen James, Pieter Abbeel, Jitendra Malik, and Trevor Darrell.
\newblock Real-world robot learning with masked visual pre-training.
\newblock In {\em Conference on Robot Learning}, pages 416--426. PMLR, 2023.

\bibitem{jing2023exploring}
Ya~Jing, Xuelin Zhu, Xingbin Liu, Qie Sima, Taozheng Yang, Yunhai Feng, and Tao Kong.
\newblock Exploring visual pre-training for robot manipulation: Datasets, models and methods.
\newblock In {\em 2023 IEEE/RSJ International Conference on Intelligent Robots and Systems (IROS)}, pages 11390--11395. IEEE, 2023.

\bibitem{laskin2020curl}
Michael Laskin, Aravind Srinivas, and Pieter Abbeel.
\newblock Curl: Contrastive unsupervised representations for reinforcement learning.
\newblock In {\em International conference on machine learning}, pages 5639--5650. PMLR, 2020.

\bibitem{sermanet2018time}
Pierre Sermanet, Corey Lynch, Yevgen Chebotar, Jasmine Hsu, Eric Jang, Stefan Schaal, Sergey Levine, and Google Brain.
\newblock Time-contrastive networks: Self-supervised learning from video.
\newblock In {\em 2018 IEEE international conference on robotics and automation (ICRA)}, pages 1134--1141. IEEE, 2018.

\bibitem{ha2018world}
David Ha and J{\"u}rgen Schmidhuber.
\newblock World models.
\newblock {\em arXiv preprint arXiv:1803.10122}, 2018.

\bibitem{escontrela2024video}
Alejandro Escontrela, Ademi Adeniji, Wilson Yan, Ajay Jain, Xue~Bin Peng, Ken Goldberg, Youngwoon Lee, Danijar Hafner, and Pieter Abbeel.
\newblock Video prediction models as rewards for reinforcement learning.
\newblock {\em Advances in Neural Information Processing Systems}, 36, 2024.

\bibitem{du2024learning}
Yilun Du, Sherry Yang, Bo~Dai, Hanjun Dai, Ofir Nachum, Josh Tenenbaum, Dale Schuurmans, and Pieter Abbeel.
\newblock Learning universal policies via text-guided video generation.
\newblock {\em Advances in Neural Information Processing Systems}, 36, 2024.

\bibitem{du2023video}
Yilun Du, Mengjiao Yang, Pete Florence, Fei Xia, Ayzaan Wahid, Brian Ichter, Pierre Sermanet, Tianhe Yu, Pieter Abbeel, Joshua~B Tenenbaum, et~al.
\newblock Video language planning.
\newblock {\em arXiv preprint arXiv:2310.10625}, 2023.

\bibitem{finn2017deep}
Chelsea Finn and Sergey Levine.
\newblock Deep visual foresight for planning robot motion.
\newblock In {\em 2017 IEEE International Conference on Robotics and Automation (ICRA)}, pages 2786--2793. IEEE, 2017.

\bibitem{gupta2022maskvit}
Agrim Gupta, Stephen Tian, Yunzhi Zhang, Jiajun Wu, Roberto Mart{\'\i}n-Mart{\'\i}n, and Li~Fei-Fei.
\newblock {MaskViT}: Masked visual pre-training for video prediction.
\newblock {\em arXiv preprint arXiv:2206.11894}, 2022.

\bibitem{baker2022video}
Bowen Baker, Ilge Akkaya, Peter Zhokov, Joost Huizinga, Jie Tang, Adrien Ecoffet, Brandon Houghton, Raul Sampedro, and Jeff Clune.
\newblock Video pretraining (vpt): Learning to act by watching unlabeled online videos.
\newblock {\em Advances in Neural Information Processing Systems}, 35:24639--24654, 2022.

\bibitem{lin2024spawnnet}
Xingyu Lin, John So, Sashwat Mahalingam, Fangchen Liu, and Pieter Abbeel.
\newblock {SpawnNet}: Learning generalizable visuomotor skills from pre-trained network.
\newblock In {\em 2024 IEEE International Conference on Robotics and Automation (ICRA)}, pages 4781--4787. IEEE, 2024.

\end{thebibliography}
\bibliographystyle{unsrt}

\end{document}